\documentclass[twoside,11pt]{article}

\usepackage{jmlr2e}

\usepackage[T1]{fontenc}
\usepackage{graphicx}

\usepackage{subfig}
\usepackage{subcaption}
\usepackage{subfloat}

\usepackage{hyperref}
\usepackage{amssymb}
\usepackage{mathtools}
\usepackage{bbold}
\usepackage{cases}
\usepackage{empheq}
\usepackage{float}
\usepackage{bbm}
\usepackage{mathrsfs}
\usepackage{multirow}
\usepackage{array}
\usepackage{physics}
\usepackage{xspace}
\usepackage[table]{xcolor}
\usepackage{xfrac}
\usepackage{tikz}
\usepackage{pgfplots}
\usepackage{url}

\usepackage{enumitem}  
\usepackage{calc}      

\usepackage{calc}      
\usetikzlibrary{calc} 
\usepackage{pgf}

\usepackage{array}
\usepackage{tabularx}

\usepackage{adjustbox}

\newtheorem{assumption}[theorem]{Assumption}

\def\app#1#2{%
  \mathrel{%
    \setbox0=\hbox{$#1\sim$}%
    \setbox2=\hbox{%
      \rlap{\hbox{$#1\propto$}}%
      \lower1.1\ht0\box0%
    }%
    \raise0.25\ht2\box2%
  }%
}
\def\approxprop{\mathpalette\app\relax}

\newcommand{\ie}{\emph{i.e.}}
\newcommand{\eg}{\emph{e.g.}}
\newcommand{\etc}{etc}

\hypersetup{citebordercolor=green}
\hypersetup{linkbordercolor=green}

\graphicspath{{./}{./plots/}}

\makeatletter
\def\input@path{{./}{./tikz/}}
\makeatother

\usepackage{lastpage}
\jmlrheading{**}{2025}{1-\pageref{LastPage}}{MM/DD; Revised MM/DD}{MM/DD}{00-0000}{R\'ois\'in Luo et al.}

\ShortHeadings{Optimization-Induced Dynamics of Lipschitz Continuity in Neural Networks}{R\'ois\'in Luo et al.}

\firstpageno{1}

\begin{document}

\title{Optimization-Induced Dynamics of Lipschitz Continuity in Neural Networks}

\author{\name R\'ois\'in Luo\thanks{Corresponding author. } \email roisincrtai@gmail.com\\
       \addr University of Galway, Ireland \\
             Irish National Centre for Research Training in AI (CRT-AI)
       \AND
       \name James McDermott \\
       \addr University of Galway, Ireland \\
             Irish National Centre for Research Training in AI (CRT-AI)
       \AND
       Christian Gagn\'e \\
       \addr Universit\'e Laval, Canada \\
             Canada-CIFAR AI Chair \\
             Mila - Qu\'ebec AI Institute
       \AND
        Qiang Sun \\ 
        \addr University of Toronto, Canada \\
              MBZUAI, UAE
        \AND
        Colm O'Riordan \\ 
        \addr University of Galway, Ireland \\
        Irish National Centre for Research Training in AI (CRT-AI)
}

\editor{My editor}

\maketitle

\begin{abstract}

Lipschitz continuity characterizes the worst-case sensitivity of neural networks to small input perturbations; yet its dynamics (\ie~temporal evolution) during training remains under-explored. We present a rigorous mathematical framework to model the temporal evolution of Lipschitz continuity during training with stochastic gradient descent (SGD). This framework leverages a system of stochastic differential equations (SDEs) to capture both deterministic force (\ie~gradient expectations) and stochastic force (\ie~gradient noise). Our theoretical analysis identifies three principal factors driving the evolution: (i) the projection of gradient flows, induced by the optimization dynamics, onto the operator-norm Jacobian of parameter matrices; (ii) the projection of gradient noise, arising from the randomness in mini-batch sampling, onto the operator-norm Jacobian; and (iii) the projection of the gradient noise onto the operator-norm Hessian of parameter matrices. Furthermore, our theoretical framework sheds light on such as how noisy supervision, parameter initialization, batch size, and mini‐batch sampling trajectories, among other factors, shape the evolution of the Lipschitz continuity of neural networks. Our experimental results demonstrate strong agreement between the theoretical implications and the observed behaviors. 




\end{abstract}

\vspace{5pt}

\section{Introduction}


Recent advancements in deep learning have led to models that excel across a broad range of domains, from vision to language. Their vulnerability to input perturbations remains a significant challenge for establishing trustworthy learning systems \citep{szegedy2013intriguing,goodfellow2014explaining,mkadry2017towards}. Lipschitz continuity (Definition~\ref{def:lipschitz_definition}) measures the worst-case sensitivity of the output of a network to small input perturbations. Furthermore, several fundamental properties, including robustness to perturbation \citep{luo2024interpreting} and generalization capability, are closely linked to network Lipschitz continuity \citep{shalev2014understanding,bartlett2017spectrally,yin2019rademacher,zhang2021understanding}. Networks with lower Lipschitz constants tend to be more resilient to input perturbations (\eg~adversarial perturbations) and exhibit improved generalization capabilities \citep{zhang2022rethinking,fazlyab2023certified,khromov2024some}.

\begin{definition}[Globally $K$-Lipschitz Continuous \citep{tao2006analysis,yosida2012functional}]
\label{def:lipschitz_definition}
Let $f: X \mapsto Y$ be a function, where $X \subseteq \mathbb{R}^d$ and $Y \subseteq \mathbb{R}^c$. The function $f$ is said to be \emph{globally $K$-Lipschitz continuous} if there exists a constant $K > 0$ such that:
\begin{equation}
    \|f(u) - f(v)\|_2 \leq K \|u - v\|_2, \quad \forall\, u, v \in X
\end{equation}
upper-bounds the function $f$.
\end{definition}

\begin{figure}[t]

  \centering
  
  \input{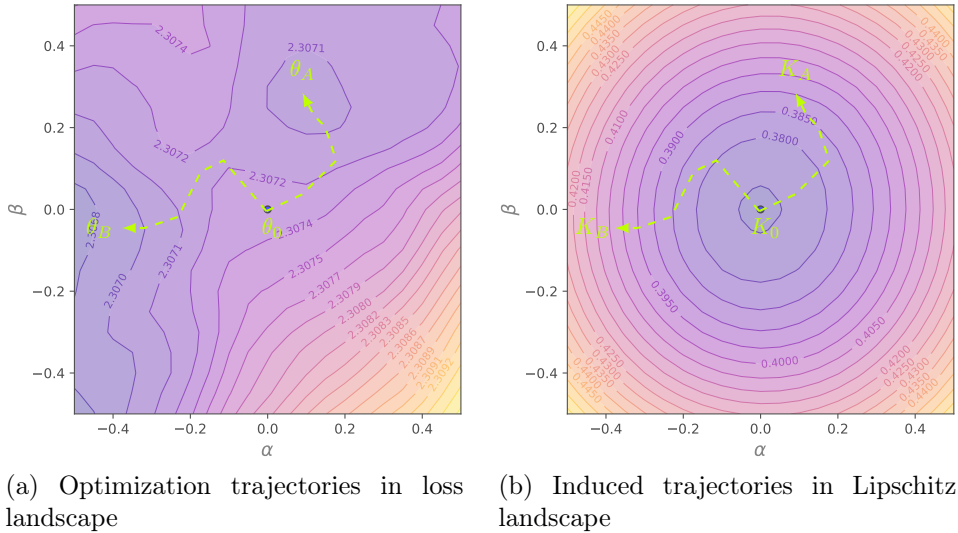}
  
  \caption[Optimization-Induced Stochastic Dynamics of Lipschitz Continuity]{\textbf{Optimization-induced dynamics}. During the training, the network parameters, starting from $\boldsymbol{\theta}_0$, moves towards a solution $\boldsymbol{\theta}_A$ or $\boldsymbol{\theta}_B$ as shown in the loss landscape~\subref{subfig:loss_landscape}, driven by optimization process. Accordingly, this dynamics, driven by the optimization, induces the evolution of the network Lipschitz continuity, starting from $K_0$ to $K_A$ or $K_B$, as shown in the Lipschitz landscape~\subref{subfig:lipschitz_landscape}. The trajectories in the loss landscape~\subref{subfig:loss_landscape} and the Lipschitz landscape~\subref{subfig:lipschitz_landscape} are visualized on the same parameter space $\alpha O \beta$. The $\alpha$ and $\beta$ are two randomly-chosen orthogonal directions in the parameter space. }

  \label{fig:lipschitz_and_loss_landscape}

\end{figure}

Gradient-based optimization methods, such as stochastic gradient descent (SGD) and its variants \citep{robbins1951stochastic,hinton2012neural,diederik2014adam}, are fundamental to training deep learning models by iteratively minimizing the loss function through parameter updates with gradients. The optimization dynamics in a neural network induce the corresponding dynamics of its Lipschitz continuity, which we refer to as \textbf{optimization‐induced dynamics}. As illustrated in Figure~\ref{fig:lipschitz_and_loss_landscape}, every optimization trajectory in the loss landscape induces a corresponding trajectory in the Lipschitz landscape. Although the dynamics (\ie~temporal evolution) of SGD has been extensively explored in the literature, including topics such as: (i) continuous‐time/SDE modeling \citep{li2019stochastic,malladi2022sdes,welling2011bayesian,zhu2018anisotropic}; (ii) edge-of-stability and implicit bias \citep{li2021happens,damian2022self,xing2018walk,jastrzkebski2017three}; (iii) convergence analysis \citep{li2017convergence}; (iv) sharp versus flat minima and generalization \citep{keskar2016large,chaudhari2019entropy,zhang2021understanding}, a comprehensive understanding of how the Lipschitz constant evolves over time during training remains lacking.

\begin{figure*}[!t]
  \centering

  \begin{minipage}[t]{1\textwidth}
     \centering 
     \subfloat[Numerical validation on CIFAR-10]{
      \includegraphics[width=1\linewidth]{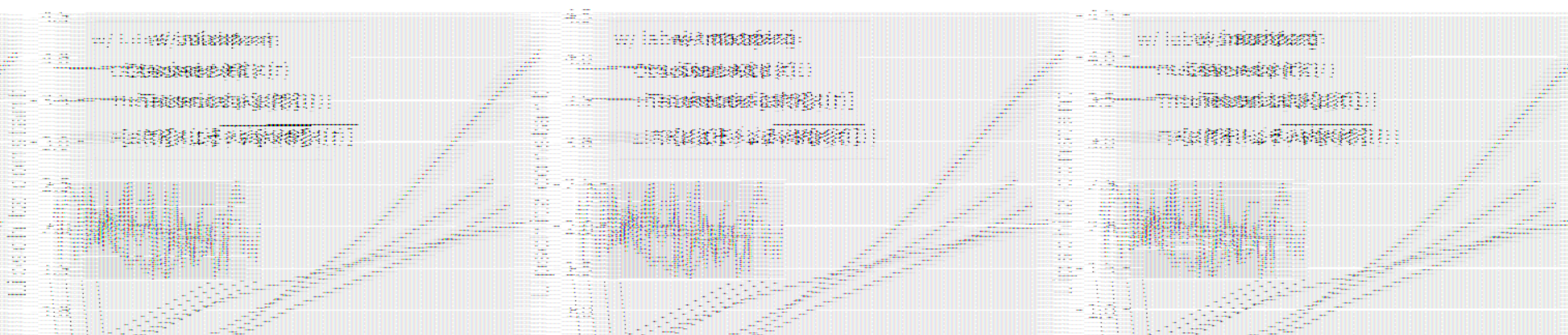}  
      \label{fig:validation:cifar10}
     }
  \end{minipage}
  
  \begin{minipage}[t]{1\textwidth}
     \centering 
     \subfloat[Numerical validation on CIFAR-100]{
      \includegraphics[width=1\linewidth]{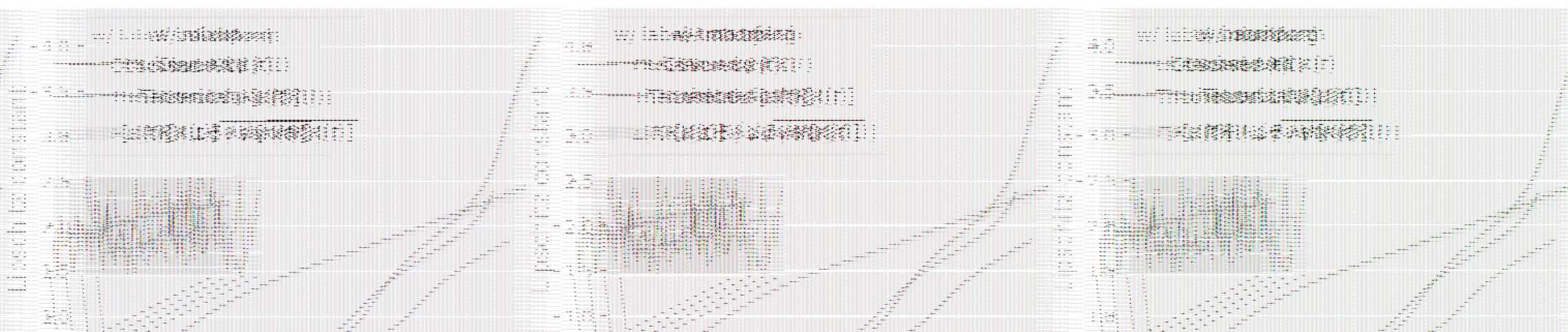}  
      \label{fig:validation:cifar100}
     }
  \end{minipage}

  \caption[Numerical Validation of Our Mathematical Framework in Characterizing Optimization-Induced Stochastic Dynamics of Lipschitz Continuity]{\textbf{Numerical validation of our mathematical framework}. The theoretical Lipschitz constants computed using our framework closely agree with empirical observations. To validate our framework, we train a five-layer ConvNet on CIFAR-10 and CIFAR-100 across multiple configurations for $30,000$ steps ($200$ epochs). We collect the instance-wise gradients over time for all layers. Using Theorem~\ref{theorem:layer_dynamics}, Theorem~\ref{theorem:network_dynamics}, Theorem~\ref{theorem:integral_form_network_dynamics} and Theorem~\ref{theorem:statistics_of_lipschitz}, we are able to theoretically compute the predicted Lipschitz continuity. The inset plots zoom in on the first $50$ steps, and demonstrate that \textbf{the trends of Lipschitz constants do not necessarily grow monotonically}. Results with more regularization configurations on CIFAR-10 are provided in Appendix~\ref{sec:full_validation_cifar10}.
  }
  
  \label{fig:validation}

\end{figure*}

This gap motivates us to establish a rigorous theoretical framework for modeling the dynamics of Lipschitz continuity, induced by optimization dynamics. Equipped with the toolkit from stochastic differential equations (SDEs) \citep{applebaum2009levy,karatzas2012brownian,oksendal2013stochastic} and operator perturbation theory \citep{kato2012short,kato2013perturbation,luo2025spectralvariations}, we can rigorously analyze the evolution of the matrix operator norm (\ie~matrix Lipschitz constant), induced by the optimization process, within an SDE framework. Our framework can capture both the deterministic and stochastic forces driving the temporal evolution of Lipschitz constants. The effectiveness of our mathematical framework is empirically verified in Figure~\ref{fig:validation}, where the theoretical implications derived using our framework (Theorem~\ref{theorem:integral_form_network_dynamics} and Theorem~\ref{theorem:statistics_of_lipschitz}) are closely aligned with the observations. Furthermore, our theoretical framework sheds light on questions such as how batch size, parameter initialization, mini‐batch sampling trajectories, label noise, \etc., shape the Lipschitz continuity evolution during optimization.

\subsection{Contributions}
\label{sec:contributions}

We highlight the key contributions below:
\begin{enumerate}

    \item \textbf{Theoretical Framework} (Section~\ref{sec:preliminaries}-\ref{sec:theoretical_analysis}). We present a rigorous mathematical framework that models the dynamics of Lipschitz continuity, leveraging an SDE system driven by Wiener processes. This SDE-based framework captures both deterministic and stochastic forces, providing a comprehensive understanding of the dynamics of Lipschitz continuity. To ensure practical applicability, we also develop a low-rank approximation method for modeling gradient noise in Section~\ref{sec:gradient_noise}, enabling efficient computation of these dynamics. 

    \item \textbf{Principal Driving Forces} (Section~\ref{sec:layer_dyanmics}-\ref{sec:statistics}). Our theoretical analysis (Theorem~\ref{theorem:integral_form_network_dynamics}) identifies three principal forces governing the dynamics: (i) the projection of gradient flows, induced by the optimization dynamics, onto the operator-norm Jacobian of parameter matrices; (ii) the projection of gradient noise, arising from the randomness in mini-batch sampling, onto the operator-norm Jacobian; and (iii) the projection of the gradient noise onto the operator-norm Hessian of parameter matrices. In the evolution of Lipschitz continuity, forces (i) and (iii) act as deterministic forces, while force (ii) modulates the stochasticity of the dynamics.

    \item \textbf{Framework Validation and Theoretical Implications} (Section~\ref{sec:validation}-\ref{sec:theoretical_implications}). Firstly, we validate our theoretical framework under multiple configurations, including \textit{batch normalization}, \textit{dropout}, \textit{weight decay}, \textit{mixup}, \textit{auto-augment}, \textit{label smoothing} and \textit{adversarial training}. Furthermore, we test the theoretical implications derived from our framework, as detailed in Section~\ref{sec:theoretical_implications}, including parameter initialization, noisy gradient regularization, uniform label corruption, batch size, and mini-batch sampling trajectories shape the evolution of Lipschitz continuity. The results show a strong agreement between the theoretical implications and observed behaviors.

\end{enumerate}

\section{Preliminaries}
\label{sec:preliminaries}

We define the notation used in our theoretical analysis. Let $f: \mathbb{R}^d \to \mathbb{R}^c$ be a function. For a time-dependent function $g$, we use $g_k$ to denote its value at discrete time step $k$, and $g(t)$ for its value at continuous time $t$. For a random variable $\xi$, \eg, representing data sampling, gradient noise, Wiener process, and filtration \citep{oksendal2013stochastic}, we consistently use subscripts, such as $\xi_k$ or $\xi_t$, for brevity. A function written as $g^{(\ell)}$ indicates that $g$ is the $\ell$-th layer of a neural network. We use $\mathrm{I}_n$ to denote a $n \times n$ identity matrix; $\mathbb{1}_n$ to denote an $n$-dimensional all-ones vector.

\medskip
\noindent
\textbf{Mini-Batch Sampling}. Let $\mathcal{D} := \{(x_i, y_i)\}_{i=1}^{N} \subseteq \mathcal{X} \times \mathcal{Y}$ denote a dataset consisting of $N$ samples, where $(x_i, y_i)$ represents the $i$-th data point and its corresponding target. Let $\xi_t := \{(x_{t_j}, y_{t_j})\}_{j=1}^{M} \subset \mathcal{X} \times \mathcal{Y}$ denote a mini-batch sampled from $\mathcal{D}$ at time point $t$, where $M$ ($M \ll N$) is the size of the mini-batch. The sequence $\{\xi_i\}_{i=0}^t$ up to time $t$ is referred to as a \textit{sampling trajectory} from $D$. The $\sigma$-algebra $\mathcal{F}_t = \sigma\bigl\{\xi_0,\xi_1,\dots,\xi_t\bigr\}$, defined on mini-batch sampling, is referred to as the \textit{filtration} generated by the sampling trajectory $\{\xi_i\}_{i=0}^t$ \citep{oksendal2013stochastic}. In stochastic analysis, $\mathcal{F}_t$ represents that the information accumulates up to time $t$ (\ie~history). Any history up to a time point $t_k$ contains the history up to a time point $t_l$ for all $t_k \geq t_l$, \ie~$\mathcal{F}_{t_l} \subseteq \mathcal{F}_{t_k}$.

\medskip
\noindent
\textbf{Feed-Forward Network}. Let $\boldsymbol{\theta}^{(\ell)} \in \mathbb{R}^{m_{\ell} \times n_{\ell}}$ denote the parameter matrix of the $\ell$-th layer of an $L$-layer feed-forward network. Let $\boldsymbol{\theta} := \{ \boldsymbol{\theta}^{(1)}, \boldsymbol{\theta}^{(2)}, \dots, \boldsymbol{\theta}^{(L)} \} \in \Theta$ denote the collection of all $L$ parameter matrices. Let $f_{\boldsymbol{\theta}}: \mathbb{R}^d \mapsto \mathbb{R}^c$ represent a feed-forward network parameterized by $\boldsymbol{\theta}$. Let $\mathrm{vec}(\boldsymbol{\theta}^{(\ell)}) \in \mathbb{R}^{m_{\ell} n_{\ell}}$ denote the vectorized $\boldsymbol{\theta}^{(\ell)}$ with \textit{column-major convention} \citep{horn2012matrix}.

\begin{remark}
The explicit use of the $\mathrm{vec}(\cdot)$ operator is essential for rigorous spectral analysis of parameter matrices beyond the usual context of parameter updates in SGD optimization.
\end{remark}

\medskip
\noindent
\textbf{Instance and Population Loss}. Let $\ell_f(\boldsymbol{\theta}; x, y): \Theta \times \mathcal{X} \times \mathcal{Y} \mapsto \mathbb{R}$ denote the instance loss for $f_{\boldsymbol{\theta}}$. Let $\mathcal{L}_f(\boldsymbol{\theta}; \xi)$ denote the population loss over a mini-batch $\xi$:
\begin{align}
    \mathcal{L}_f(\boldsymbol{\theta}; \xi) := \frac{1}{M} \sum_{x_{i}, y_{i} \in \xi} \ell_f(\boldsymbol{\theta}; x_{i}, y_{i}) \notag
    .
\end{align}
Let $\mathcal{L}_f(\boldsymbol{\theta})$ denote the population loss over the dataset $\mathcal{D}$:
\begin{align}
    \mathcal{L}_f(\boldsymbol{\theta}) := \mathbb{E}_{(x_i, y_i) \sim \mathcal{D}} \left[ \ell_f(\boldsymbol{\theta}; x_i, y_i) \right] \notag
    .
\end{align}

\medskip
\noindent
\textbf{Unbiased Gradient Estimator}. For brevity, we use:
\begin{align}
\nabla^{(\ell)} \mathcal{L}_f(\boldsymbol{\theta}):=\nabla_{\boldsymbol{\theta}^{(\ell)}} \mathcal{L}_f(\boldsymbol{\theta})   , \quad \text{and} \quad 
\nabla \mathcal{L}_f(\boldsymbol{\theta}) := \nabla_{\boldsymbol{\theta}} \mathcal{L}_f(\boldsymbol{\theta}) \notag
, 
\end{align}
to denote the gradient with respect to the $\ell$-th layer parameter matrix, and the gradient with respect to the collective parameter matrices, respectively. Note that $\nabla \mathcal{L}_f(\boldsymbol{\theta}; \xi)$ is an unbiased gradient estimator for $\nabla \mathcal{L}_f(\boldsymbol{\theta})$:
\begin{align}
    \mathbb{E} \left[ \nabla L_t(\boldsymbol{\theta}; \xi) \right] = \mathbb{E} \left[ \nabla \mathcal{L}_f(\boldsymbol{\theta}) \right] = \mathbb{E} \left[ \nabla \ell_f(\boldsymbol{\theta}; x, y) \right] \notag
    .
\end{align}

\section{Vectorized SDE for Continuous-Time SGD}

SGD and its variants (\eg~Adam) \citep{robbins1951stochastic,hinton2012neural,diederik2014adam} serve as cornerstone optimization algorithms widely used for training deep neural networks. Formally, at a time point $k$, the SGD-based update regarding the parameter $\boldsymbol{\theta}_{k}^{(\ell)}$ with a mini-batch $\xi_k$ can be formulated by:
\begin{align}
    \label{equ:sgd}
    \boldsymbol{\theta}_{k+1}^{(\ell)} = \boldsymbol{\theta}_{k}^{(\ell)} - \eta \nabla^{(\ell)} \mathcal{L}_f(\boldsymbol{\theta}_k; \xi_k) 
    ,
\end{align}
where $\eta$ is the learning rate. Some studies \citep{mandt2015continuous,su2016differential,stephan2017stochastic} view SGD as a deterministic process with a deterministic ordinary differential equation (ODE) as:
\begin{align}
    \frac{\dd \boldsymbol{\theta}^{(\ell)}(t)}{\dd t} = - \nabla^{(\ell)} \mathcal{L}_f(\boldsymbol{\theta}(t)) \notag
    ,
\end{align}
where $t \approx k\eta$ and $\dd t \approx \eta \to 0$.

However, the inherent stochasticity in the population gradients $\nabla^{(\ell)} \mathcal{L}_f(\boldsymbol{\theta}_k; \xi_k)$, arising from the randomness in mini-batch sampling $\xi_k$, is not accounted for by ODE methods. The \textit{layer-wise batch gradient noise} (\ie~batch gradient fluctuations) \citep{welling2011bayesian, keskar2016large, chaudhari2019entropy,zhang2021understanding}, defined as a positive semi-definite (PSD) matrix:
\begin{align}
\label{equ:batch_gradient_noise}
 \boldsymbol{\Sigma}_k^{(\ell)} := \mathrm{Var}\left[ \mathrm{vec}\left(\nabla^{(\ell)} \mathcal{L}_f(\boldsymbol{\theta}_k; \xi_k) \right)\right]  = \frac{1}{M} \mathrm{Var}\left[ \mathrm{vec}\left(\nabla^{(\ell)} \ell_f(\boldsymbol{\theta}_k; x, y) \right)\right] 
 \in \mathbb{R}^{m_{\ell}n_{\ell} \times m_{\ell}n_{\ell}}
 ,
\end{align}
induced by mini-batch sampling, can influence the optimization trajectory. \textbf{It remains under-explored in the literature how such stochasticity affects the evolution of Lipschitz continuity of neural networks over time}.  

\begin{assumption}[Collective Gradient Noise Structure Assumption]
\label{assump:layerwise_noise}
For tractability, we assume that gradient‐noise covariances between different layers are negligible. Equivalently, $\Sigma_t$ has block‐diagonal structure:
\begin{align}
   \boldsymbol{\Sigma}_t = 
   \mathrm{diag}(\boldsymbol{\Sigma}_t^{(1)}, \boldsymbol{\Sigma}_t^{(2)}, \cdots, \boldsymbol{\Sigma}_t^{(L)})  \notag
   ,
\end{align}
so that we model batch gradient noise on a per‐layer basis only. This layer‐wise approximation is common in large‐scale SDE analyses, \eg~\citep{grosse2016kronecker,malladi2022sdes,simsekli2019tail}.
\end{assumption}

To address the limitations of ODE methods, SDE methods \citep{li2017convergence,jastrzkebski2017three,zhu2018anisotropic,chaudhari2019entropy} extend ODE methods by accounting for gradient noise. SDE-based methods capture the stochasticity inherent in mini-batch updates and its effects on the optimization trajectory. The multivariate Central Limit Theorem (CLT) states that the population gradient estimator:
\begin{align}
 \nabla^{(\ell)} \mathcal{L}_f(\boldsymbol{\theta}_k; \xi_k) = \frac{1}{M} \sum_{(x_i, y_i) \in \xi_k} \nabla^{(\ell)} \ell_f(\boldsymbol{\theta}_k; x_i, y_i) \notag
 ,
\end{align}
distributionally converges to a normal distribution in $\mathbb{R}^{m_{\ell}n_{\ell}}$:
\begin{align}
\mathrm{vec}\left(\nabla^{(\ell)} \mathcal{L}_f(\boldsymbol{\theta}_k; \xi_k)\right)  \xrightarrow{\text{d}}  \mathcal{N}\left(\mathrm{vec}(\nabla^{(\ell)} \mathcal{L}_f(\boldsymbol{\theta}_k)), \boldsymbol{\Sigma}_k^{(\ell)}\right) \notag
,
\end{align}
as $M \to \infty$, where the samples $(x_i, y_i)$ are \emph{i.i.d}. Therefore, SGD can be modeled as an It\^o process, which provides a more accurate representation of the dynamics by considering the continuous-time SDEs \citep{mandt2015continuous, stephan2017stochastic}.

\begin{definition}[Vectorized SDE for Continuous-Time SGD]
Under Assumption~\ref{assump:layerwise_noise}, and the assumption that $\nabla^{(\ell)} \mathcal{L}_f(\boldsymbol{\theta}_t)$ and $\boldsymbol{\Sigma}_t^{(\ell)}$ satisfy global Lipschitz and linear‐growth conditions \citep{oksendal2013stochastic,karatzas2012brownian}, the SGD update for the parameter $\boldsymbol{\theta}_t^{(\ell)}$ with a mini-batch $\xi_t$ at time $t$ can be formulated as a matrix-valued It\^o's SDE by:
\begin{align}
\label{equ:ct_sgd}
\dd  \mathrm{vec}  \left(\boldsymbol{\theta}^{(\ell)}(t)\right) &= -\mathrm{vec} \left[\nabla^{(\ell)} \mathcal{L}_f(\boldsymbol{\theta}(t))\right] \dd t + \sqrt{\eta}\left[\boldsymbol{\Sigma}_t^{(\ell)} \right]^{\frac{1}{2}}\dd \boldsymbol{B}_t^{(\ell)}
,
\end{align}
where the batch gradient noise $\left[\boldsymbol{\Sigma}_t^{(\ell)}\right]^{\sfrac{1}{2}} \in \mathbb{R}^{m_{\ell}n_{\ell} \times m_{\ell}n_{\ell}}$ satisfies:
\begin{align}
   \left(\boldsymbol{\Sigma}_t^{(\ell)}\right)^{\frac{1}{2}}
   \left[\left(\boldsymbol{\Sigma}_t^{(\ell)}\right)^{\frac{1}{2}}\right]^\top
    = \boldsymbol{\Sigma}_t^{(\ell)},
    \quad \text{and} \quad 
    \mathbb{R}^{m_{\ell}n_{\ell}} \ni \dd \boldsymbol{B}_t^{(\ell)} \sim \mathcal{N}(\boldsymbol{0}, \boldsymbol{\mathrm{I}}_{m_{\ell}n_{\ell}} \dd t) \notag
\end{align}
represents the infinitesimal increment of a Wiener process (standard Brownian motion) adapted to the filtration $\mathcal{F}_t$ in $\mathbb{R}^{m_{\ell}n_{\ell}}$. 

\label{def:ct_sgd}
\end{definition}

\section{Estimating batch gradient noise}
\label{sec:gradient_noise}

The covariance of gradient noise structure reflects how neurons interact with each other. Capturing the full structure of gradient noise for numerically analyzing the spectra of network parameters \citep{stewart1990matrix,horn2012matrix,kato2013perturbation} is shaped by the stochasticity arising from mini batch sampling. However, modeling the full gradient noise covariance is computationally prohibitive: accurately estimating batch gradient noise requires sampling all gradient trajectories over the entire dataset, which leads to intractable storage and computational costs \citep{mandt2015continuous,li2019stochastic,chaudhari2018stochastic}.

SDE-based models often have strong assumptions regarding the structure of gradient noise. For example, in the literature \citep{welling2011bayesian,stephan2017stochastic,jastrzkebski2017three}, the gradient noise is reduced into a constant scalar. Under this assumption, the SDE reduces to an Ornstein-Uhlenbeck process \citep{oksendal2013stochastic,karatzas2012brownian}. However, this simplification overlooks the covariance structures inherent in mini-batch sampling. To address this, some literature assumes a diagonal structure for the gradient noise \citep{jastrzkebski2018relation,simsekli2020fractional}. While this approximation captures varying variances across parameters, it neglects potential correlations between them.

In practice, \citeauthor{mandt2015continuous} compute the exact $2\times2$ covariance matrix for a low‐dimensional logistic‐regression problem;  \citeauthor{zhu2018anisotropic} estimate the full noise covariance --- solely to extract its leading eigenpairs --- in an MLP with several hundred parameters \citep{zhu2018anisotropic}. Nonetheless, these methods do not scale to modern deep networks, where the number of parameters can be on the order of millions. In the remaining part of this section, we aim to develop experiment-friendly method for tracking full structure of gradient noise.

\subsection{Unbiased batch gradient noise estimator}

To capture the complex interactions among neurons, we model the complete structure of the gradient noise. However, the computational cost of $\boldsymbol{\Sigma}_t^{(\ell)}$, which requires the evaluation on the entire dataset $\mathcal{D}$ for each $t$, poses a challenge to the applicability of our framework. This computation remains highly demanding, even when using state-of-the-art GPUs. To overcome this, we develop a low-rank approximation method. We now seek to estimate $\boldsymbol{\Sigma}_t^{(\ell)}$ without bias. 

\begin{proposition}[Unbiased Batch Gradient Noise Estimator]
\label{prop:unbiased_batch_gradient_noise_estimator}
Starting from Equation~\ref{equ:batch_gradient_noise}, the batch gradient noise at time $t$ is estimated by:
\begin{align}
\label{equ:unbiased_batch_gradient_noise_estimator}
\boldsymbol{\Sigma}_k^{(\ell)} 
&\approx \frac{1}{M} \left[\frac{1}{M-1} 
\sum_{x_{t_i},y_{t_i} \in \xi_t}
\underbrace{
\left\{ \mathrm{vec}\left[\nabla^{(\ell)} \ell_f(\boldsymbol{\theta}(t); x_{t_i}, y_{t_i})  - \nabla^{(\ell)} \mathcal{L}_f(\boldsymbol{\theta}(t);\xi_t)\right]\right\}
}_{(\boldsymbol{\Omega}_{t_i}^{(\ell)})^\top}
\right.\notag\\
&\left.\qquad\qquad\qquad\qquad\qquad
\underbrace{
\left\{ \mathrm{vec}\left[\nabla^{(\ell)} \ell_f(\boldsymbol{\theta}(t); x_{t_i}, y_{t_i})  - \nabla^{(\ell)} \mathcal{L}_f(\boldsymbol{\theta}(t);\xi_t)\right]\right\}^\top
}_{\boldsymbol{\Omega}_{t_i}^{(\ell)}}
\right] \notag\\
&=\frac{1}{M}
\left(
\frac{1}{\sqrt{M-1}}
\begin{bmatrix}
    \boldsymbol{\Omega}_{t_1}^{(\ell)} \\
    \boldsymbol{\Omega}_{t_2}^{(\ell)} \\
    \vdots\\
    \boldsymbol{\Omega}_{t_M}^{(\ell)}
\end{bmatrix}
\right)^\top
\left(
\frac{1}{\sqrt{M-1}}
\begin{bmatrix}
    \boldsymbol{\Omega}_{t_1}^{(\ell)}\\
    \boldsymbol{\Omega}_{t_2}^{(\ell)}\\
    \vdots\\
    \boldsymbol{\Omega}_{t_M}^{(\ell)}
\end{bmatrix}
\right)=\frac{1}{M} (\boldsymbol{\Omega}_t^{(\ell)})^\top \boldsymbol{\Omega}_t^{(\ell)}
,
\end{align}
where $\boldsymbol{\Omega}_{t_i}^{(\ell)}$ is the point-wise gradient fluctuation:
\begin{align}
(\boldsymbol{\Omega}_{t_i}^{(\ell)})^\top = \mathrm{vec}\left[\nabla^{(\ell)} \ell_f(\boldsymbol{\theta}(t); x_{t_i}, y_{t_i})  - \nabla^{(\ell)} \mathcal{L}_f(\boldsymbol{\theta}(t);\xi_t)\right]    \in \mathbb{R}^{m_{\ell}n_{\ell}} \notag
,
\end{align}
and $\boldsymbol{\Omega}_t^{(\ell)}$ is batch-wise gradient fluctuation:
\begin{align} 
\boldsymbol{\Omega}_t^{(\ell)} &=
\frac{1}{\sqrt{M-1}}
\begin{bmatrix}
    \boldsymbol{\Omega}_{t_1}^{(\ell)}\\
    \boldsymbol{\Omega}_{t_2}^{(\ell)}\\
    \vdots\\
    \boldsymbol{\Omega}_{t_M}^{(\ell)}
\end{bmatrix}
=
\sqrt{\frac{1}{M-1}}
    \begin{pmatrix}
    \mathrm{vec}\left[\nabla^{(\ell)} \ell_f(\boldsymbol{\theta}(t); x_{t_1}, y_{t_1}) - \nabla^{(\ell)} \mathcal{L}_f(\boldsymbol{\theta}(t);\xi_t)\right]^\top \\
    \mathrm{vec}\left[\nabla^{(\ell)} \ell_f(\boldsymbol{\theta}(t); x_{t_2}, y_{t_2}) - \nabla^{(\ell)} \mathcal{L}_f(\boldsymbol{\theta}(t);\xi_t)\right]^\top \\
    \vdots \\
    \mathrm{vec}\left[\nabla^{(\ell)} \ell_f(\boldsymbol{\theta}(t); x_{t_M}, y_{t_M}) - \nabla^{(\ell)} \mathcal{L}_f(\boldsymbol{\theta}(t);\xi_t)\right]^\top
    \end{pmatrix} \notag
    .
\end{align}

\end{proposition}

\subsection{Estimating variance and covariance}

To study the contributions of variances (diagonal elements) and covariances (off-diagonal elements) of gradient noise to the dynamics, we decompose the gradient noise into variance and covariance components using Proposition~\ref{prop:variance_and_covariance_of_gradient_noise}.

\begin{proposition}[Diagonal and Off-Diagonal Elements of Batch Gradient Noise]
\label{prop:variance_and_covariance_of_gradient_noise}
The diagonal variance $\boldsymbol{\Sigma}_{t}^{(\ell)}\mid_{\mathrm{var}}$ and the off-diagonal covariance $\boldsymbol{\Sigma}_{t}^{(\ell)}\mid_{\mathrm{cov}}$ are computed, respectively, by:
\begin{align}
     \boldsymbol{\Sigma}_{t}^{(\ell)}\mid_{\mathrm{var}} \approx \frac{1}{M}\left(\boldsymbol{\Omega}_t^{(\ell)} \odot \boldsymbol{\Omega}_t^{(\ell)}\right)^\top \boldsymbol{\mathbb{1}}_{M} 
     ,
     \quad 
     \text{and}
     \quad
      \boldsymbol{\Sigma}_{t}^{(\ell)}\mid_{\mathrm{cov}} \approx \boldsymbol{\Sigma}_t^{(\ell)} - \boldsymbol{\Sigma}_{t}^{(\ell)}\mid_{\mathrm{var}} \notag
      ,
\end{align}
where $\odot$ denotes Hadamard product. Note that $\boldsymbol{\Sigma}_{t}^{(\ell)}\mid_{\mathrm{cov}}$ is not necessarily a PSD matrix. 
\end{proposition}

\begin{proof}
Starting from Equation~\ref{equ:unbiased_batch_gradient_noise_estimator}, this can be obtained by:
    \begin{align}
    \left(\boldsymbol{\Sigma}_{t}^{(\ell)}\mid_{\text{var}} \right)_{i,i} &=\left(\Sigma_t^{(\ell)}\right)_{i,i} 
        \approx \frac{1}{M} \left( (\boldsymbol{\Omega}_t^{(\ell)})^\top \boldsymbol{\Omega}_t^{(\ell)} \right)_{i,i} \notag
        = \frac{1}{M}\sum_{j=1}^{M} \left((\boldsymbol{\Omega}_t^{(\ell)})^\top\right)_{i,j} \boldsymbol{\Omega}_{j,i}^{(\ell)} \notag\\
        &=  \frac{1}{M}\sum_{j=1}^{M} \boldsymbol{\Omega}_{j,i}^{(\ell)} \boldsymbol{\Omega}_{j,i}^{(\ell)} 
        = \frac{1}{M}\sum_{j=1}^{M} \left(\boldsymbol{\Omega}_t^{(\ell)} \odot \boldsymbol{\Omega}_t^{(\ell)} \right)_{j,i} 
        = \frac{1}{M}\left(\left(\boldsymbol{\Omega}_t^{(\ell)} \odot \boldsymbol{\Omega}_t^{(\ell)} \right)^\top \mathbb{1}_{M}\right)_i \notag
        .
    \end{align}
\end{proof}

\subsection{Computing square root}

Direct computation of $(\boldsymbol{\Sigma}_t^{(\ell)})^{\sfrac{1}{2}}$ by
\begin{align}
(\boldsymbol{\Sigma}_t^{(\ell)})^{\frac{1}{2}} 
\approx
\left[
    \frac{1}{M} (\boldsymbol{\Omega}_t^{(\ell)})^\top \boldsymbol{\Omega}_t^{(\ell)}
    \right]^{\frac{1}{2}}
    \in
    \mathbb{R}^{m_{\ell}n_{\ell} \times m_{\ell}n_{\ell}} \notag
\end{align}
is computationally infeasible for large parameter matrices. Note that
\begin{align}
 \frac{1}{M} \boldsymbol{\Omega}_t^{(\ell)}(\boldsymbol{\Omega}_t^{(\ell)})^\top   \in \mathbb{R}^{M \times M}
 ,
 \quad
 M \ll m_{\ell}n_{\ell} \notag
 ,
\end{align}
has low-rank structure and the same spectrum of $\boldsymbol{\Sigma}_t^{(\ell)}$. We seek an efficient method by exploiting the low-rank SVD in $\frac{1}{M}\boldsymbol{\Omega}_t^{(\ell)}(\boldsymbol{\Omega}_t^{(\ell)})^\top$.

\begin{proposition}[Square Root Approximation of Covariance Matrix]
\label{prop:}

Suppose $\sfrac{\boldsymbol{\Omega}_t^{(\ell)}}{\sqrt{M}}$ admits SVD:
\begin{align}
\frac{\boldsymbol{\Omega}_t^{(\ell)}}{\sqrt{M}} = \boldsymbol{U}_t^{(\ell)} (\boldsymbol{\Lambda}_t^{(\ell)})^{\frac{1}{2}} (\boldsymbol{V}_t^{(\ell)})^\top \notag
, 
\end{align}
where $\boldsymbol{\Lambda}_t^{(\ell)}$ and $\boldsymbol{U}_t^{(\ell)}$ are the eigenvalues and eigenvectors of $\frac{1}{M} \boldsymbol{\Omega}_t^{(\ell)}(\boldsymbol{\Omega}_t^{(\ell)})^\top$. Then:
\begin{align}
(\boldsymbol{\Sigma}_t^{(\ell)})^{\frac{1}{2}}    
\approx
\left[
\frac{ (\boldsymbol{\Omega}_t^{(\ell)})^\top \boldsymbol{U}_t^{(\ell)}}{\sqrt{M}}
\right]
(\boldsymbol{\Lambda}_t^{(\ell)})^{-\frac{1}{2}}
\left[
\frac{ (\boldsymbol{\Omega}_t^{(\ell)})^\top \boldsymbol{U}_t^{(\ell)}}{\sqrt{M}}
\right]^\top \notag
.
\end{align}
\end{proposition}

\begin{proof}
Starting from Equation~\ref{equ:unbiased_batch_gradient_noise_estimator}, this can be obtained by:
\begin{align}
(\boldsymbol{\Sigma}_t^{(\ell)})^{\frac{1}{2}}
&\approx
\left[
    \frac{1}{M} (\boldsymbol{\Omega}_t^{(\ell)})^\top \boldsymbol{\Omega}_t^{(\ell)}
\right]^{\frac{1}{2}}
=
\left[
\frac{(\boldsymbol{\Omega}_t^{(\ell)})^\top}
{\sqrt{M}}
\boldsymbol{U}_t^{(\ell)} (\boldsymbol{U}_t^{(\ell)})^\top
\frac{\boldsymbol{\Omega}_t^{(\ell)}}{\sqrt{M}} 
\right]^{\frac{1}{2}}
\notag\\
&=
\left[
\frac{(\boldsymbol{\Omega}_t^{(\ell)})^\top\,\boldsymbol{U}_t^{(\ell)} }{\sqrt{M}}
(\boldsymbol{\Lambda}_t^{(\ell)})^{-\frac{1}{2}}(\boldsymbol{\Lambda}_t^{(\ell)})
(\boldsymbol{\Lambda}_t^{(\ell)})^{-\frac{1}{2}}
\frac{(\boldsymbol{U}_t^{(\ell)})^\top\,\boldsymbol{\Omega}_t^{(\ell)}}{\sqrt{M}} 
\right]^{\frac{1}{2}} 
\end{align}

Set:
\begin{align}
\boldsymbol{P}=\frac{(\boldsymbol{\Omega}_t^{(\ell)})^\top\,\boldsymbol{U}_t^{(\ell)} }
{\sqrt{M}}
(\boldsymbol{\Lambda}_t^{(\ell)})^{-\frac{1}{2}} \notag
, 
\end{align}
and note:
\begin{align}
\boldsymbol{P}=
\frac{(\boldsymbol{\Omega}_t^{(\ell)})^\top\,\boldsymbol{U}_t^{(\ell)} }
{\sqrt{M}}
(\boldsymbol{\Lambda}_t^{(\ell)})^{-\frac{1}{2}}
 =  \left[
\boldsymbol{U}_t^{(\ell)} (\boldsymbol{\Lambda}_t^{(\ell)})^{\frac{1}{2}} (\boldsymbol{V}_t^{(\ell)})^\top 
 \right]^\top 
 \boldsymbol{U}_t^{(\ell)} (\boldsymbol{\Lambda}_t^{(\ell)})^{-\frac{1}{2}}  = \boldsymbol{V}_t^{(\ell)} \notag
\end{align}
is orthonormal.

Hence:
\begin{align}
(\boldsymbol{\Sigma}_t^{(\ell)})^{\frac{1}{2}}
&\approx
\left[
\frac{(\boldsymbol{\Omega}_t^{(\ell)})^\top\,\boldsymbol{U}_t^{(\ell)} }{\sqrt{M}}
(\boldsymbol{\Lambda}_t^{(\ell)})^{-\frac{1}{2}}(\boldsymbol{\Lambda}_t^{(\ell)})
(\boldsymbol{\Lambda}_t^{(\ell)})^{-\frac{1}{2}}
\frac{(\boldsymbol{U}_t^{(\ell)})^\top\,\boldsymbol{\Omega}_t^{(\ell)}}{\sqrt{M}} 
\right]^{\frac{1}{2}} \notag\\
&=
\left[
\frac{(\boldsymbol{\Omega}_t^{(\ell)})^\top\,\boldsymbol{U}_t^{(\ell)} }{\sqrt{M}}
(\boldsymbol{\Lambda}_t^{(\ell)})^{-\frac{1}{2}}
\right]
(\boldsymbol{\Lambda}_t^{(\ell)})^{\frac{1}{2}}
\left[
(\boldsymbol{\Lambda}_t^{(\ell)})^{-\frac{1}{2}}
\frac{(\boldsymbol{U}_t^{(\ell)})^\top\,\boldsymbol{\Omega}_t^{(\ell)}}{\sqrt{M}} 
\right]^{\frac{1}{2}} \notag\\
&=
\underbrace{
\left[
\frac{(\boldsymbol{\Omega}_t^{(\ell)})^\top\,\boldsymbol{U}_t^{(\ell)} }
{\sqrt{M}}
(\boldsymbol{\Lambda}_t^{(\ell)})^{-\frac{1}{2}}
\right]
}_{\boldsymbol{P}}
(\boldsymbol{\Lambda}_t^{(\ell)})^{\frac{1}{2}}
\underbrace{
\left[
\frac{(\boldsymbol{\Omega}_t^{(\ell)})^\top\,\boldsymbol{U}_t^{(\ell)} }
{\sqrt{M}}
(\boldsymbol{\Lambda}_t^{(\ell)})^{-\frac{1}{2}}
\right]^\top
}_{\boldsymbol{P}^\top}
\notag\\
&=\left[
\frac{(\boldsymbol{\Omega}_t^{(\ell)})^\top\,\boldsymbol{U}_t^{(\ell)} }
{\sqrt{M}}
\right]
(\boldsymbol{\Lambda}_t^{(\ell)})^{-\frac{1}{2}}
\left[
\frac{(\boldsymbol{\Omega}_t^{(\ell)})^\top\,\boldsymbol{U}_t^{(\ell)} }
{\sqrt{M}}
\right]^\top \notag
.
\end{align}


\end{proof}

\section{Temporal evolution of Lipschitz upper bound}


Suppose that an $L$-layer feed-forward neural network $f$ is the composition of $L$ layers, each comprising a linear unit with an activation unit, expressed as:
\begin{align}
    f := \left(\rho^{(L)} \circ~\phi^{(L)}\right) \circ \cdots \circ~\left(\rho^{(2)} \circ~\phi^{(2)} \right)\circ~\left(\rho^{(1)} \circ~\phi^{(1)}\right) \nonumber
\end{align}
where $\left(\rho^{(\ell)} \circ~\phi^{(\ell)}\right)$ represents the $\ell$-th layer consisting of the linear unit $\psi^{(\ell)}(\cdot)$ and the activation unit $\rho^{(\ell)}(\cdot)$. Let the Lipschitz constants of $\rho^{(\ell)}$ and $\phi^{(\ell)}$ be $A^{(\ell)}$ and $K^{(\ell)}$, respectively. The upper bound of the network Lipschitz constant of $f$ then is the product of $A^{(\ell)}$ and $K^{(\ell)}$ across all layers, as shown in Proposition~\ref{prop:network_lipschitz_bound} \citep{miyato2018spectral,fazlyab2019efficient,virmaux2018lipschitz,gouk2021regularisation,virmaux2018lipschitz}.

\begin{proposition}[Lipschitz Continuity Bound in Feed-Forward Network]
\label{prop:network_lipschitz_bound}
Let $\hat{K}(t)$ be the Lipschitz constant of a feed-forward neural network $f$ without skip connections. The network Lipschitz constant $\hat{K}(t)$ is upper-bounded by:
\begin{align}
    \hat{K}(t):=\sup_{\forall~u \neq v} \frac{\|f(u) - f(v)\|_2}{\|u-v\|_2} \leq K(t):=\prod_{l=1}^{L} A^{(\ell)} \cdot \prod_{l=1}^{L} K^{(\ell)}(t) \notag
    .
\end{align} 
This upper bound is useful in robust certification and theoretical analysis of deep models with complex topologies \citep{fazlyab2019efficient, gouk2021regularisation, virmaux2018lipschitz}. We refer to $K(t)$ as Lipschitz continuity or Lipschitz constant for brevity.
\end{proposition}

Since most activation functions, such as ReLU, Leaky ReLU, Tanh, Sigmoid, \etc, have $1$-Lipschitz continuity \citep{virmaux2018lipschitz}, we set $A^{(\ell)}=1$ for brevity in theoretical analysis. Note that Proposition~\ref{prop:network_lipschitz_bound} does not take into account skip connections. 


\begin{proposition}[Operator Norm of Linear Unit] 
\label{prop:operator_norm_of_linear_unit}
Suppose the $\phi^{(\ell)}(t)$ is a linear unit with matrix multiplication or convolution:
\begin{align}
    \phi^{(\ell)}(t)(z) := \boldsymbol{\theta}^{(\ell)}(t)(z) + \boldsymbol{b}^{(\ell)}(t),
\end{align}
where $\boldsymbol{\theta}^{(\ell)}(t)$ and $\boldsymbol{b}^{(\ell)}(t)$ are the $\ell$-th layer parameter matrix and bias, respectively. Then, the operator norm of $\phi^{(\ell)}(t)$ (\ie~spectral norm) admits:
\begin{align}
\|\phi^{(\ell)}(t)\|_{op} = \|\boldsymbol{\theta}^{(\ell)}(t)\|_{op} = \sigma_{1}^{(\ell)}(t) \notag
,
\end{align}
where $\sigma_{1}^{(\ell)}(t)$ is the largest singular value of $\theta^{(\ell)}(t)$ \citep{miyato2018spectral,sedghi2019singular,yoshida2017spectral}.
\end{proposition}

\begin{definition}[Stochastic Dynamical System of Lipschitz Continuity Bound]
\label{def:lipschitz_dynamical_system}
Collect: 
\begin{enumerate}
    \item Definition~\ref{def:ct_sgd}: Vectorized SDE for Continuous-Time SGD;
    \item Proposition~\ref{prop:network_lipschitz_bound}: Lipschitz Continuity Bound in Feed-Forward Network;
    \item Proposition~\ref{prop:operator_norm_of_linear_unit}: Operator Norm of Linear Unit,
\end{enumerate}
so that the dynamics of Lipschitz continuity bound for a feed-forward neural network is characterized by a system of SDEs:
\begin{numcases}{}
    \dd \mathrm{vec}(\boldsymbol{\theta}^{(\ell)}(t)) = - \mathrm{vec}\left[\nabla^{(\ell)} \mathcal{L}_f(\boldsymbol{\theta}(t))\right]  \, \dd t   + \sqrt{\eta} \left[\boldsymbol{\Sigma}_t^{(\ell)} \right]^{\frac{1}{2}} \, \dd \boldsymbol{B}_t^{(\ell)} \notag\\
    K^{(\ell)}(t)=\|\boldsymbol{\theta}^{(\ell)}(t)\|_{op}  \notag\\
    Z(t)  = \sum_{l=1}^L \log K^{(\ell)}(t) \notag\\
    K(t) = \mathrm{e}^{Z(t)} \notag
\end{numcases}
adapted to the filtration $\mathcal{F}_t$, where: (i) $K^{(\ell)}(t)$ governs the $\ell$-th layer dynamics of Lipschitz continuity; (ii) $K(t)$ governs the network dynamics of Lipschitz continuity bound; and (iii) $Z(t)$ is the logarithmic Lipschitz constant to facilitate theoretical deductions. \textbf{This stochastic dynamical system characterizes the dynamics of the Lipschitz continuity bound induced by the optimization dynamics}.

\end{definition}

\begin{figure*}[t!]
    \centering
    
  \begin{minipage}[t]{1\textwidth}
     \centering 
     \subfloat[Layer-specific dynamics]{
      \includegraphics[width=1\linewidth]{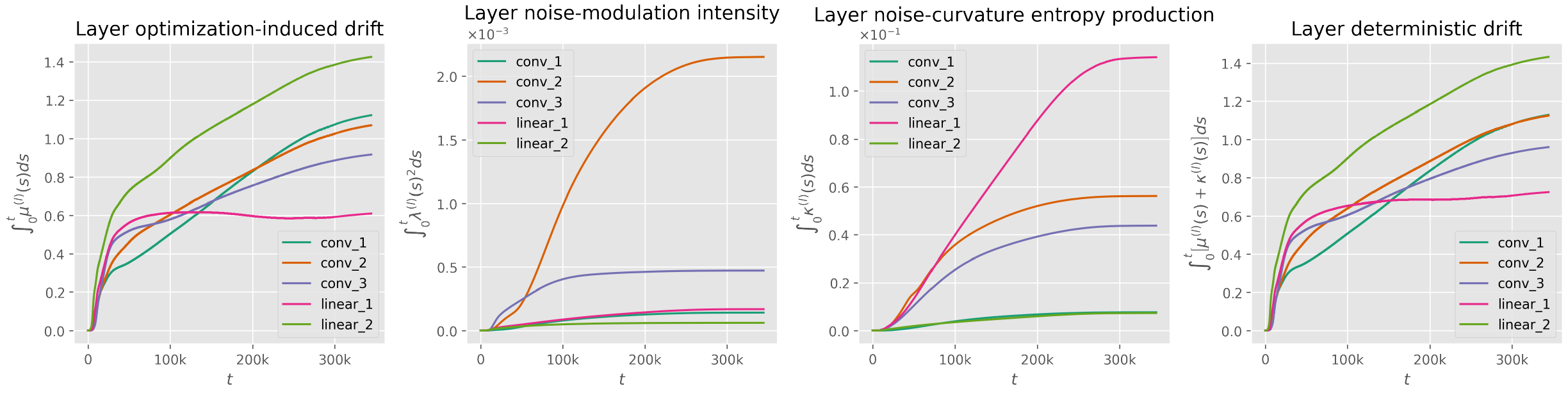}  
      \label{fig:dynamics:layer}
     }
  \end{minipage}
  \vfill
  \begin{minipage}[t]{1\textwidth}
      \centering
      \subfloat[Network-specific dynamics (differential)]{
      \includegraphics[width=1\linewidth]{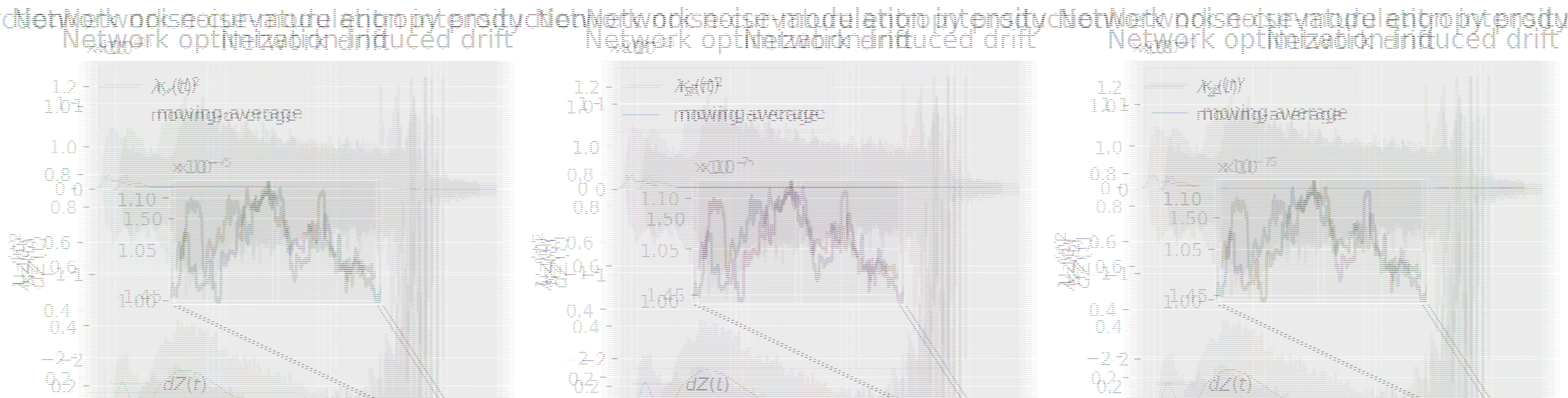}  
      \label{fig:dynamics:network:differential}
     }
  \end{minipage}
  \vfill
  \begin{minipage}[t]{1\textwidth}
      \centering
      \subfloat[Network-specific dynamics (integral)]{
      \includegraphics[width=1\linewidth]{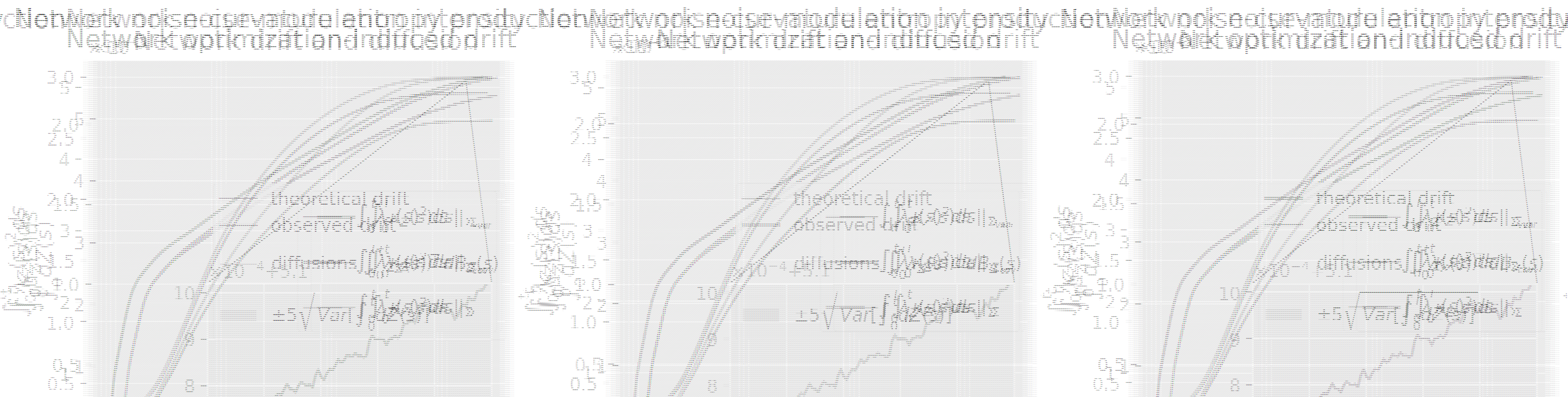}  
      \label{fig:dynamics:network}
     }
  \end{minipage}


   
  \caption[Stochastic Dynamics of Lipschitz Continuity Near Convergence]{Dynamics near convergence. We profile both layer-specific and network-specific dynamics over $344,370$ steps ($1766$ epochs) on CIFAR-10. At the end of training, the final training loss and test loss are $9.75 \times 10^{-3}$ and $2.22$, respectively; the final training accuracy and test accuracy are $0.99996$ and $0.68540$, respectively. To investigate how the variances (\ie~diagonal elements) and covariances (\ie~off‐diagonal elements) of the gradient noise affect the dynamics, the dynamics are computed with respect to variances ($\Sigma_{\mathrm{var}}$) and covariances ($\Sigma_{\mathrm{cov}}$) respectively, using Theorem~\ref{theorem:network_dynamics}. The results indicate that: (i) the optimization‐trajectory drift plays the primary role in shaping Lipschitz continuity over time; (ii) the covariances dominate the noise contributions; and (iii) the noise-curvature entropy production $\kappa_Z(t)$ remains significant near convergence, leading to a gradual and steady increase in Lipschitz continuity. The inset plot zooms in on $100$ steps at $t=330,000$. The moving averages are computed over a window of $500$ steps.}
  
  \label{fig:longrange_dynamics}
  
\end{figure*}

\section{Theoretical analysis}
\label{sec:theoretical_analysis}

Note that the equation system in Definition~\ref{def:lipschitz_dynamical_system} is driven by a Wiener process, which consists of stochastic differential equations (SDEs), and is adapted to the filtration $\mathcal{F}_t$, with time-dependent parameters $\theta^{(\ell)}(t)$. We leverage It\^o's Lemma \citep{ito1951stochastic,oksendal2013stochastic} to analyze this SDE system. We aim to analyze the dynamics of $Z(t)$ and $K(t)$. The sketch for theoretical analysis is:
\begin{enumerate} 

    \item \textbf{Layer-Specific Dynamics} (Section~\ref{sec:layer_dyanmics}). We derive the SDE for $\dd K_t^{(\ell)}$ by applying It\^o's Lemma to the process $\dd \mathrm{vec}\left[\theta^{(\ell)}(t)\right]$. This step establishes a direct connection between parameter-level updates driven by the optimization process and the layer-wise operator norm.
      
    \item \textbf{Network-Specific Dynamics} (Section~\ref{sec:network_dynamics}). Building on the layer-specific analysis, we derive the network-level SDE for $\dd Z(t)$ by applying It\^o's Lemma to the processes $\dd K^{(\ell)}(t)$ across $L$ layers. The process $Z(t)$ captures the logarithmic, network-level dynamics of Lipschitz continuity. Finally, we obtain the dynamics of $K(t)$ from $\dd Z(t)$ via It\^o calculus.
        
    \item \textbf{Statistical Characterization} (Section~\ref{sec:statistics}). We analyze the statistical properties of $Z(t)$ and $K(t)$, including their expectations and variances. These results provide insight into the asymptotic behavior of the system near convergence.
\end{enumerate}

\subsection{First- and second-order operator-norm derivatives}
\label{sec:jacobian_and_hessian}

Stochastic spectral analysis regarding the parameter matrices requires the first-order and second-order derivatives of the largest singular values with respect to parameter matrices. The first-order operator-norm derivative is well-known in the literature \citep{luo2025spectralvariations,kato2013perturbation,horn2012matrix,magnus2019matrix} (Lemma~\ref{lemma:jacobian_spectral_norm}). The second-order operator-norm derivative can be derived through perturbation theory \citep{kato2013perturbation,luo2025spectralvariations}. Lemma~\ref{lemma:hessian_spectral_norm} is deducted in the literature \citep{luo2025spectralvariations} based on perturbation theory for linear operators \citep{kato2012short,kato2013perturbation}. Let $\boldsymbol{\theta}^{(\ell)}(t)$ admit a SVD:
\begin{align}
    \boldsymbol{\theta}_t^{(\ell)} = \sum_{i=1}^{r} 
    \sigma_{i}^{(\ell)}(t) \boldsymbol{u}_{i}^{(\ell)} (t)
    \boldsymbol{v}_{i}^{(\ell)}(t)^\top
    ,\quad
    \sigma_{1}^{(\ell)}(t) > \sigma_{2}^{(\ell)}(t) > \cdots > \sigma_{r}^{(\ell)}(t)
    ,\quad
    r = \rank(\boldsymbol{\theta}^{(\ell)}(t))
    ,\notag
\end{align}
under the Assumption~\ref{assump:differentiability}.

\begin{assumption}[\textbf{Spectral Differentiability}]
\label{assump:differentiability}
We assume the singular values of parameter matrices are \emph{simple}: $\sigma_i^{(\ell)} \neq \sigma_j^{(\ell)}$, for all $i \neq j$. This assumption guarantees $\sigma_i^{(\ell)}, \boldsymbol{u}_i^{(\ell)}, \boldsymbol{v}_i^{(\ell)} \in C^{\infty}$ (\ie~differentiable at arbitrary integer order) \citep{kato2013perturbation,kato2012short}.
\end{assumption}

\begin{lemma}[Operator-Norm Jacobian]
\label{lemma:jacobian_spectral_norm}
The operator-norm Jacobian of $\|\boldsymbol{\theta}^{(\ell)}(t)\|_{op}$ with respect to $\mathrm{vec}\left[\boldsymbol{\theta}^{(\ell)}(t)\right]$ is given by:
\begin{align}
\boldsymbol{J}_{op}^{(\ell)}(t) 
=\frac{\partial \|\boldsymbol{\theta}^{(\ell)}(t)\|_{op}}{\partial \operatorname{vec}\left[\boldsymbol{\theta}^{(\ell)}(t)\right]}
=\boldsymbol{v}_{1}^{(\ell)}(t) \otimes \boldsymbol{u}_{1}^{(\ell)}(t) 
\in \mathbb{R}^{m_{\ell}n_{\ell}}
\notag
,
\end{align} 
where $\otimes$ denotes Kronecker product (\ie~tensor product) \citep{luo2025spectralvariations,kato2013perturbation,horn2012matrix,magnus2019matrix}.
\end{lemma}

\begin{lemma}[Operator-Norm Hessian \citep{luo2025spectralvariations}]
\label{lemma:hessian_spectral_norm}
The operator-norm Hessian of $\|\boldsymbol{\theta}^{(\ell)}(t)\|_{op}$ with respect to $\mathrm{vec}\left[\boldsymbol{\theta}^{(\ell)}(t)\right]$ is given by:
\begin{align}
\boldsymbol{H}_{op}^{(\ell)}(t) 
=
\frac{\partial}{\partial \operatorname{vec}\left[ \boldsymbol{\theta}^{(\ell)}(t)\|_{op} \right]}
\operatorname{vec}\left[\frac{\partial \|\boldsymbol{\theta}^{(\ell)}(t)\|_{op}}{\partial \boldsymbol{\theta}^{(\ell)}(t)}\right]
= \boldsymbol{H}_{L}^{(\ell)}(t) + \boldsymbol{H}_{R}^{(\ell)}(t) + \boldsymbol{H}_{C}^{(\ell)}(t) 
\in \mathbb{R}^{m_{\ell}n_{\ell} \times m_{\ell}n_{\ell}}
\notag
,
\end{align}
where:
\begin{gather}
\boldsymbol{e}_{i,j}^{(\ell)}(t) := \boldsymbol{v}^{(\ell)}_i(t) \otimes \boldsymbol{u}^{(\ell)}_j(t) , \notag \\
\boldsymbol{H}_{L}^{(\ell)}(t) =\sum_{i \neq 1, i \leq m_{\ell}} \frac{\sigma_1^{(\ell)}(t)}{\sigma_1^{(\ell)}(t)^2 - \sigma_i^{(\ell)}(t)^2} \boldsymbol{e}_{1,i}^{(\ell)}(t) \otimes  \boldsymbol{e}_{1,i}^{(\ell)}(t)^\top  ,\notag\\
\boldsymbol{H}_{R}^{(\ell)}(t) = \sum_{j \neq 1, j \leq n_{\ell}} \frac{\sigma_1^{(\ell)}(t)}{\sigma_1^{(\ell)}(t)^2 - \sigma_j^{(\ell)}(t)^2} \boldsymbol{e}_{j,1}^{(\ell)}(t) \otimes \boldsymbol{e}_{j,1}^{(\ell)}(t)^\top , \notag\\
\boldsymbol{H}_{C}^{(\ell)}(t) = \sum_{k \neq 1, k \leq r_{\ell}} \frac{\sigma_k^{(\ell)}(t)}{\sigma_1^{(\ell)}(t)^2 - \sigma_k^{(\ell)}(t)^2} 
\left[
\boldsymbol{e}_{1,k}^{(\ell)}(t) \otimes \boldsymbol{e}_{k,1}^{(\ell)}(t)^\top +  \boldsymbol{e}_{k,1}^{(\ell)}(t) \otimes \boldsymbol{e}_{1,k}^{(\ell)}(t)^\top \right] \notag 
,
\end{gather}
and $r_{\ell}=\rank(\boldsymbol{\theta}^{(\ell)}(t))$.

\end{lemma}

\begin{remark}
Note that the function $\sigma_1^{(\ell)}(t): \boldsymbol{\theta}^{(\ell)}(t) \mapsto \mathbb{R}$ is convex. Since the Hessian of a convex function is PSD, it follows that $\boldsymbol{H}_{\text{op}}^{(\ell)}(t)$ is PSD.
\end{remark}

\subsection{Layer-specific dynamics}
\label{sec:layer_dyanmics}

\begin{theorem}[Layer-Specific Dynamics]
\label{theorem:layer_dynamics}
The layer-specific dynamics of Lipschitz continuity is given by:
\begin{align}
\frac{\dd K^{(\ell)}(t)}{K^{(\ell)}(t)}  = \left( \mu^{(\ell)}(t) + \kappa^{(\ell)}(t) \right) \,  \dd t + \boldsymbol{\lambda}^{(\ell)}(t)^\top \, \dd \boldsymbol{B}_t^{(\ell)} 
, 
\end{align}    
where $\dd\,\boldsymbol{B}_t^{(\ell)} \sim \mathcal{N}(\boldsymbol{0}, \boldsymbol{\mathrm{I}}_{m_{\ell}n_{\ell}} \dd t)$ represents the increment of a standard Wiener process in $\mathbb{R}^{m_{\ell}n_{\ell}}$, and:
\begin{gather}
\mu^{(\ell)}(t) =  \frac{1}{\sigma_{1}^{(\ell)}(t)} \left\langle \boldsymbol{J}_{op}^{(\ell)}(t), \, -\mathrm{vec}\left[ \nabla^{(\ell)} \mathcal{L}_f(\boldsymbol{\theta}(t)) \right] \right\rangle ,\notag \\
\boldsymbol{\lambda}^{(\ell)}(t) = \frac{\sqrt{\eta}}{\sigma_{1}^{(\ell)}(t)}  
\left(\left[\boldsymbol{\Sigma}_t^{(\ell)}\right]^{\frac{1}{2}}\right)^\top 
\boldsymbol{J}_{op}^{(\ell)}(t)
, \quad
\|\boldsymbol{\lambda}^{(\ell)}(t)\|_2^2 = \boldsymbol{\lambda}^{(\ell)}(t)^\top \boldsymbol{\lambda}^{(\ell)}(t) \geq 0,\notag \\
\kappa^{(\ell)}(t) =\frac{\eta}{2\sigma_{1}^{(\ell)}(t)} \left\langle \boldsymbol{H}_{op}^{(\ell)}(t),  \boldsymbol{\Sigma}_t^{(\ell)}\right\rangle \geq 0,\notag 
\end{gather}
where we refer to:
\begin{enumerate}

    \item $\mu^{(\ell)}(t)$ as \textbf{layer optimization-induced drift}, representing the contribution of the gradient flow $\nabla^{(\ell)} \mathcal{L}_f(\boldsymbol{\theta}(t))$ --- induced by the optimization process --- to the expectation of Lipschitz continuity. This term corresponds to the projection of the negative gradient expectation onto the principal subspace (\ie~largest singular value) of the parameter matrix, and acts as a deterministic drift in the evolution of the Lipschitz constant.

    \item $\boldsymbol{\lambda}^{(\ell)}(t)$ as \textbf{layer diffusion-modulation intensity}, representing the contribution of the gradient noise $\boldsymbol{\Sigma}_t^{(\ell)}$ --- arising from the randomness in mini-batch sampling --- to the stochasticity of Lipschitz continuity. This term modulates the stochasticity of Lipschitz continuity and governs the uncertainty of temporal evolution.

    \item $\kappa^{(\ell)}(t)$ as \textbf{layer noise-curvature entropy production}, representing the non-negative, irreversible contribution of the gradient noise $\boldsymbol{\Sigma}_t^{(\ell)}$ to the deterministicity of Lipschitz continuity. Intuitively, the dynamical system baths in the stochastic gradient fluctuations captured by $\boldsymbol{\Sigma}_t^{(\ell)}$, arising from mini-batch sampling. The stochastic fluctuations ``dissipate'' into the system via the curvature of the operator-norm landscape $\boldsymbol{H}_{op}^{(\ell)}(t)$, driving an irreversible increase in the Lipschitz constant and entropy.
    
\end{enumerate}    

\end{theorem}


\begin{proof}
Start with Definition~\ref{def:lipschitz_dynamical_system}:
\begin{numcases}{}
\dd  \mathrm{vec}  \left(\boldsymbol{\theta}^{(\ell)}(t)\right) = -\mathrm{vec} \left[\nabla^{(\ell)} \mathcal{L}_f(\boldsymbol{\theta}(t))\right] \dd t + \sqrt{\eta}\left[\boldsymbol{\Sigma}_t^{(\ell)} \right]^{\frac{1}{2}}\dd \boldsymbol{B}_t^{(\ell)} \label{equ:layer_proof_ct_sgd}\\
K^{(\ell)}(t) = \|\boldsymbol{\theta}^{(\ell)}(t)\|_{op} = \sigma_1^{(\ell)}(t) \label{equ:layer_proof_operator_norm}
\end{numcases}

Using Lemma~\ref{lemma:jacobian_spectral_norm} and Lemma~\ref{lemma:hessian_spectral_norm}, we apply It\^o's Lemma on Equation~\ref{equ:layer_proof_operator_norm}:
\begin{align}
    \dd K^{(\ell)}(t) = \underbrace{
    \boldsymbol{J}_{op}^{(\ell)}(t)^\top \dd \operatorname{vec}\left(\boldsymbol{\theta}^{(\ell)}(t)\right)
    }_{A} + 
    \underbrace{
    \frac{1}{2} \left(\dd \operatorname{vec}(\boldsymbol{\theta}^{(\ell)}(t)\right)^\top \; \boldsymbol{H}_{op}^{(\ell)}(t) \; \dd \operatorname{vec}(\boldsymbol{\theta}^{(\ell)}(t))}_{B}.
\end{align}

\medskip
\noindent
\textbf{Compute $A$ and $B$}.
\begin{enumerate}
    \item Substitute Equation~\ref{equ:layer_proof_ct_sgd} into $A$:
\begin{align}
\label{equ:layer_proof_expanded_A}
    A &= \boldsymbol{J}_{op}^{(\ell)}(t)^\top \dd \operatorname{vec}\left(\boldsymbol{\theta}^{(\ell)}(t)\right) \notag\\
    &=\boldsymbol{J}_{op}^{(\ell)}(t)^\top \left[ - \mathrm{vec}\left[\nabla^{(\ell)} \mathcal{L}_f(\boldsymbol{\theta}(t))\right] \, \dd t   + \sqrt{\eta} \left[\boldsymbol{\Sigma}_t^{(\ell)}\right]^{\frac{1}{2}} \, \dd \boldsymbol{B}_t^{(\ell)} \right] \notag\\
    &= \left\langle \boldsymbol{J}_{op}^{(\ell)}(t), - \mathrm{vec}\left[\nabla^{(\ell)} \mathcal{L}_f(\boldsymbol{\theta}(t))\right] \right\rangle \dd t + \boldsymbol{J}_{op}^{(\ell)}(t)^\top \sqrt{\eta} \left[\boldsymbol{\Sigma}_t^{(\ell)}\right]^{\frac{1}{2}} \, \dd \boldsymbol{B}_t^{(\ell)}. 
\end{align}

\item Use trace identity $\tr(XYZ)=\tr(YZX)$:
\begin{align}
    B &= \frac{1}{2} \left(\dd \operatorname{vec}(\boldsymbol{\theta}^{(\ell)}(t)\right)^\top \boldsymbol{H}_{op}^{(\ell)}(t)  \dd \operatorname{vec}(\boldsymbol{\theta}^{(\ell)}(t)) \notag\\
    &= \frac{1}{2} \tr\left[  \boldsymbol{H}_{op}^{(\ell)}(t)  \, \dd \operatorname{vec}(\boldsymbol{\theta}^{(\ell)}(t))  \left(\dd \operatorname{vec}(\boldsymbol{\theta}^{(\ell)}(t)\right)^\top \right]. \notag
\end{align}

Substituting Equation~\ref{equ:layer_proof_ct_sgd} into $B$, we obtain:
\begin{align}
   B &= \frac{1}{2} \tr\left[ \boldsymbol{H}_{op}^{(\ell)}(t) \dd \operatorname{vec}(\boldsymbol{\theta}^{(\ell)}(t)) \left(\dd \operatorname{vec}(\boldsymbol{\theta}^{(\ell)}(t)\right)^\top \right] \notag\\
   &= \frac{1}{2} \tr\left[ \boldsymbol{H}_{op}^{(\ell)}(t) \eta \left[\boldsymbol{\Sigma}_t^{(\ell)}\right]^{\frac{1}{2}} \left\{\left[\boldsymbol{\Sigma}_t^{(\ell)}\right]^{\frac{1}{2}}\right\}^\top \dd \boldsymbol{B}_t^{(\ell)} (\dd \boldsymbol{B}_t^{(\ell)})^\top \right]\notag \\
    &=\frac{1}{2}\eta \tr\left[ \boldsymbol{H}_{op}^{(\ell)}(t)\left[\boldsymbol{\Sigma}_t^{(\ell)}\right]^{\frac{1}{2}} \left\{\left[\boldsymbol{\Sigma}_t^{(\ell)}\right]^{\frac{1}{2}}\right\}^\top  \right] \dd t \notag \\
    &=\frac{1}{2} \eta \tr\left[  \boldsymbol{H}_{op}^{(\ell)}(t)   \boldsymbol{\Sigma}_t^{(\ell)} \right] \dd t 
    =
    \frac{1}{2} \eta \left\langle  \boldsymbol{H}_{op}^{(\ell)}(t),   \boldsymbol{\Sigma}_t^{(\ell)}\right\rangle \dd t,  
\end{align}
by dropping higher-order infinitesimal terms, as $ o\left((\dd t)^2\right) \to 0$ and $o\left(\dd t \dd \boldsymbol{B}_t^{(\ell)}\right) \to \boldsymbol{0}$.
 
\end{enumerate}

\medskip
\noindent
\textbf{Combine $A+B$}. Combine expanded $A$ and $B$:
\begin{align}
    \frac{\dd K^{(\ell)}(t)}{K^{(\ell)}(t)} 
    &= \left[
    \underbrace{\frac{1}{\sigma_1^{(\ell)}(t)} \left\langle \boldsymbol{J}_{op}^{(\ell)}(t), - \mathrm{vec}\left[\nabla^{(\ell)} \mathcal{L}_f(\boldsymbol{\theta}(t))\right] \right\rangle
    }_{ \mu^{(\ell)}(t) }+ 
    \underbrace{
    \frac{\eta}{2\sigma_1^{(\ell)}(t)}  \left\langle  \boldsymbol{H}_{op}^{(\ell)}(t),   \boldsymbol{\Sigma}_t^{(\ell)}\right\rangle 
    }_{ \kappa^{(\ell)}(t) }
    \right]\dd t \notag \\
    & \qquad +
    \underbrace{
    \frac{\sqrt{\eta}}{\sigma_1^{(\ell)}(t)} \boldsymbol{J}_{op}^{(\ell)}(t)^\top  \left[\boldsymbol{\Sigma}_t^{(\ell)}\right]^{\frac{1}{2}}
    }_{ \boldsymbol{\lambda}^{(\ell)}(t)^\top }
    \, \dd \boldsymbol{B}_t^{(\ell)}.
\end{align}

\end{proof}

\subsection{Network-specific dynamics}
\label{sec:network_dynamics}


\begin{theorem}[Logarithmic Network-Specific Dynamics]
\label{theorem:network_dynamics}
The logarithmic, network-specific dynamics of Lipschitz continuity is given by:
\begin{align}
\label{equ:log_dynamics}
   \dd Z(t) = \left( \mu_Z(t) + \kappa_Z(t) - \frac{1}{2}  \lambda_Z(t)^2 \right) \dd t + \lambda_Z(t) \, \dd W_t,  
\end{align}
where $\dd W_t \sim \mathcal{N}(0, \dd t)$ represents the increment of a standard Wiener process  adapted to the filtration $\mathcal{F}_t$ in $\mathbb{R}$ by:
\begin{gather}
    \dd W_t := 
    \left[\sum_{\ell=1}^{L} \|\boldsymbol{\lambda}^{(\ell)}(t)\|_2^2\right]^{-\frac{1}{2}}
    \cdot
    \sum_{\ell=1}^{L} \boldsymbol{\lambda}^{(\ell)}(t)^\top \, \dd \boldsymbol{B}_t^{(\ell)} \quad \sim\quad \mathcal{N}(0, \dd t), \notag\\
    \mu_Z(t) = \sum_{\ell=1}^{L} \mu^{(\ell)}(t) \in \mathbb{R} , \quad
    \kappa_Z(t) = \sum_{\ell=1}^{L}   \kappa^{(\ell)}(t) \in \mathbb{R}_{+},\quad
    \lambda_Z(t) = \left[\sum_{\ell=1}^{L} \|\boldsymbol{\lambda}^{(\ell)}(t)\|_2^2\right]^{\frac{1}{2}} \in \mathbb{R}_{+}, \notag
\end{gather}   
where we refer to: (i) $\mu_Z(t)$ as \textbf{network optimization-induced drift}; (ii) $\lambda_Z(t)$ as \textbf{network diffusion-modulation intensity}; and (iii) $\kappa_Z(t)$ as \textbf{network noise-curvature entropy production}, respectively. 
\end{theorem}

\begin{proof}
Consider the differentials:
\begin{align}
\frac{\partial Z(t)}{\partial K^{(\ell)}(t)}  = \frac{1}{K^{(\ell)}(t)} ,
\quad
\text{and}\quad 
\frac{\partial^2 Z(t)}{\partial K^{(\ell)}(t)^2}  = -\frac{1}{K^{(\ell)}(t)^2} .
\end{align}

We apply It\^o's Lemma on $\dd K^{(\ell)}$ across all layers to derive the dynamics of:
\begin{align}
    Z(t)  = \sum_{l=1}^L \log K^{(\ell)}(t) \notag
    ,
\end{align}
so that:
\begin{align}
    &\dd Z(t)  = \dd \left( \sum_{l=1}^L \log K^{(\ell)}(t) \right) = \sum_{l=1}^L \left( 
    \underbrace{
    \frac{\partial Z(t)}{\partial K^{(\ell)}(t)} \dd K^{(\ell)}(t)
    }_{C} 
    + 
    \underbrace{
    \frac{1}{2} \frac{\partial^2 Z(t)}{\partial K^{(\ell)}(t)^2} \left(\dd K^{(\ell)}(t)
    \right)^2 
    }_{D}
    \right) \notag
    .
\end{align}

\textbf{Derive $C$ and $D$}.
\begin{enumerate}
    \item Derive $C$:
    \begin{align}
    C=\frac{\partial Z(t)}{\partial K^{(\ell)}(t)} \dd K^{(\ell)}(t) = \frac{\dd K^{(\ell)}(t)}{K^{(\ell)}(t)} = \left( \mu^{(\ell)}(t) + \kappa^{(\ell)}(t) \right)  \, \dd t + \boldsymbol{\lambda}^{(\ell)}(t)^\top  \dd \boldsymbol{B}_t^{(\ell)} \notag .
\end{align}

    \item Derive $D$:
\begin{align}
   D=\frac{1}{2} \frac{\partial^2 Z(t)}{\partial K^{(\ell)}(t)^2} \left(\dd K^{(\ell)}(t)
    \right)^2 = -\frac{1}{2K^{(\ell)}(t)^2} \left(\dd K^{(\ell)}(t)
    \right)^2 = -\frac{1}{2} \|\boldsymbol{\lambda}^{(\ell)}(t)\|_2^2 \dd t \notag
    ,
\end{align}
by dropping higher-order infinitesimal terms as $o((\dd t)^2) \to 0$ and $ o(\dd t \dd \boldsymbol{B}_t^{(\ell)}) \to \boldsymbol{0}$.

\end{enumerate}

\medskip
\noindent
\textbf{Combine $C+D$}. Hence:
\begin{align}
   \dd Z(t) &= \left( \sum_{l=1}^{L} \mu^{(\ell)}(t) + \kappa^{(\ell)}(t) - \frac{1}{2}\|\boldsymbol{\lambda}^{(\ell)}(t)\|_2^2 \right) \dd t +  \sum_{\ell=1}^{L} \boldsymbol{\lambda}^{(\ell)}(t)^\top \dd \boldsymbol{B}_t^{(\ell)} \nonumber \\
&= \left( \mu_Z(t) + \kappa_Z(t) - \frac{1}{2}  \lambda_Z(t)^2 \right) \dd t + \lambda_Z(t) \dd W_t \nonumber 
\label{equ:log_dynamics}
\end{align}
where $\dd W_t \sim \mathcal{N}(0, \dd t)$ represents the increment of a Wiener process in $\mathbb{R}$.

\end{proof}

\begin{theorem}[Integral-Form Dynamics of Lipschitz Continuity]
\label{theorem:integral_form_network_dynamics}
Using It\^o's calculus, the integral-form dynamics of Lipschitz continuity is stated as:
 \begin{align}
    \begin{cases}
    Z(t)  =  Z(0) + \displaystyle\int_0^t \left[ \mu_Z(s) + \kappa_Z(s) - \frac{1}{2} \lambda_Z(s)^2 \right] \, \dd s +  \displaystyle\int_0^t \lambda_Z(s) \, \dd W_s     \\
    K(t) =\exp\left\{Z(t)\right\} \nonumber
    \end{cases}
\end{align}
where $Z(0)$ is the initial value of $Z(t)$.
\end{theorem}

\subsection{Statistical characterization}
\label{sec:statistics}

\begin{theorem}[Statistics of Lipschitz Continuity]
\label{theorem:statistics_of_lipschitz}
Let $K(0)$ be the initial Lipschitz continuity bound, and hence $Z(0) = \log K(0)$. The expectation and variance of $K(t)$ are given as:
\begin{gather}
\mathbb{E}[K(t)] = \mathrm{e}^{\mathbb{E}[Z(t)] + \frac{1}{2}\mathrm{Var}\left[Z(t)\right]} 
= K(0)\cdot 
\mathrm{e}^{\int_0^t \mu_Z(s) \dd s}
\cdot 
\mathrm{e}^{\int_0^t  \kappa_Z(s)   \dd s}
\\
\mathrm{Var}[K(t)] = \mathbb{E}[K(t)]^2 \left(\mathrm{e}^{\mathrm{Var}[Z(t)]} - 1\right)
=\mathbb{E}[K(t)]^2
\left(
\mathrm{e}^{\int_0^t  \lambda_Z(s)^2   \dd s}  - 1
\right)
,
\end{gather}
where:
\begin{gather}
    \mathbb{E}[Z(t)] = Z(0) + \int_0^t \left[ \mu_Z(s) + \kappa_Z(s) - \frac{1}{2} \lambda_Z(s)^2 \right] \dd s  
    \\
    \mathrm{Var}\left[Z(t)\right] = \int_0^t \lambda_Z(s)^2 \, \dd s .
\end{gather}
\end{theorem}

\begin{remark}


The expectation of Lipschitz continuity is dominated only by three factors: 
\begin{enumerate}[label=\roman*]
    \item \textbf{Parameter initialization $K(0)$}. The parameter initialization determines the network Lipschitz continuity. We discuss the details in Section~\ref{sec:parameter_initialization}.
    
    \item \textbf{Optimization-induced drift $\mu_Z(t)$}. The projection of gradient expectations on the operator-norm Jacobian, induced by optimization process, drive the evolution of Lipschitz continuity;
    
    \item \textbf{Noise-curvature entropy production $\kappa_Z(t)$}. The gradient fluctuations, arising from mini-batch sampling, has a deterministic, non-negative and irreversible increase in the Lipschitz continuity.
\end{enumerate}

The uncertainty of Lipschitz continuity is determined by two factors:
\begin{enumerate}[label=\roman*]
    \item \textbf{Lipschitz continuity expectation $\mathbb{E}\left[K(t)\right]$}. The uncertainty is proportional to the current Lipschitz constant bound. Larger Lipschitz constant bound leads  to larger uncertainty of the evolution;
    
    \item \textbf{Diffusion-modulation intensity $\lambda_Z(t)$}. The projection of the gradient fluctuations on the operator-norm Jacobian modulates the diffusion process of Lipschitz continuity, hence dominates the uncertainty of the evolution.
    
\end{enumerate}    
\end{remark}

\begin{proof}



\begin{lemma}[Moment-Generating Function of $Z(t)$]
\label{lemma:mgf_zt}
Consider $Z(t)$:
\begin{align}
    Z(t) \sim \mathcal{N}\left(\mathbb{E}\left[Z(t)\right], \operatorname{Var}\left[Z(t)\right]\right)
    ,
\end{align}
then the MGF of $Z(t)$ is given as:
    \begin{align}
      \mathbb{E}\left[e^{kZ(t)}\right] =e^{k \mathbb{E}\left[Z(t)\right] + \frac{1}{2}k^2\operatorname{Var}\left[Z(t)\right]}.
    \end{align}
\end{lemma}

Set $k=1$:
\begin{align}
    \mathbb{E}\left[K(t)\right]  &= \mathbb{E}\left[e^{Z(t)}\right] = e^{ \mathbb{E}\left[Z(t)\right] + \frac{1}{2}\operatorname{Var}\left[Z(t)\right]}
    = \mathrm{e}^{Z(0) + \int_0^t \left[ \mu_Z(s) + \kappa_Z(s)  \right] \dd s}
    ,
\end{align}
so that:
\begin{align}
\left(\mathbb{E}\left[K(t)\right]\right)^2  &= \left(\mathbb{E}\left[e^{Z(t)}\right]\right)^2 = e^{ 2\mathbb{E}\left[Z(t)\right] + \operatorname{Var}\left[Z(t)\right]}
=\mathrm{e}^{2Z(0) + \int_0^t \left[ 2\mu_Z(s) + 2\kappa_Z(s)  \right] \dd s}
.
\end{align}

Set $k=2$:
\begin{align}
    \mathbb{E}\left[ K(t)^2 \right]  &= \mathbb{E}\left[e^{2Z(t)}\right] = e^{ 2\mathbb{E}\left[Z(t)\right] + 2\operatorname{Var}\left[Z(t)\right]}.
\end{align}

Hence:
\begin{align}
\operatorname{Var}\left[K(t)\right] &= \mathbb{E}\left[ K(t)^2 \right] - \left(\mathbb{E}\left[K(t)\right]\right)^2 
=\mathbb{E}\left[K(t)\right]^2\left( e^{\operatorname{Var}\left[Z(t)\right]} - 1\right).
\end{align}

\end{proof}

\section{Framework validation}
\label{sec:validation}

\textbf{Experimental Settings}. We conduct numerical experiments to validate our framework, using a five-layer ConvNet (details in Table~\ref{tab:network_config_cnn} in the Appendix), optimized with SGD (without momentum). Experimental details are as follows:
\begin{enumerate}

    \item \textbf{Datasets}. To validate the effectiveness of our framework across various datasets, the ConvNet is trained on CIFAR-10 and CIFAR-100 respectively, for $39,000$ steps ($200$ epochs) with a learning rate of $10^{-3}$ and a batch size of $256$. 

    \item \textbf{Regularizers}. To validate the effectiveness of our framework under various regularization configurations (see Table~\ref{tab:regularization_config} in Appendix), the ConvNet is trained with multiple regularizers, including \textit{batch normalization}, \textit{mixup}, and \textit{label smoothing}. Additional results with \textit{dropout}, \textit{weight decay}, \textit{auto-augment}, and \textit{adversarial training} are presented in the Appendix.
       
\end{enumerate}

\medskip
\noindent
\textbf{Experimental Results}. Figure~\ref{fig:validation:cifar10} presents the results on CIFAR-10, and Figure~\ref{fig:validation:cifar100} shows the corresponding results on CIFAR-100. These experiments demonstrate that our mathematical framework closely captures the dynamics of Lipschitz continuity across a range of regularizers.

\section{Theoretical implications}
\label{sec:theoretical_implications}

Our theoretical framework provides a unified perspective for analyzing and interpreting the evolution of Lipschitz continuity in neural networks. Specifically, it enables us to answer the following questions:
\begin{enumerate}

    \item Section~\ref{sec:key_factors}: \textbf{Key factors}. What are the key factors governing the temporal evolution of Lipschitz continuity during training?

    \item Section~\ref{sec:parameter_initialization}: \textbf{Parameter initialization}. How does parameter initialization affect the temporal evolution of Lipschitz continuity?

    \item Section~\ref{sec:near_convergence}: \textbf{Unbounded growth near convergence}. What dynamics emerge as the network approaches convergence, particularly in the regime where the training loss approaches zero?

    
    \item Section~\ref{sec:noisy_gradient}: \textbf{Noisy gradient regularization}. How does gradient noise regulate the temporal evolution of Lipschitz continuity?
    
  
    \item Section~\ref{sec:uniform_label_noise}: \textbf{Uniform label corruption}. How does uniform label corruption implicitly regularize the dynamics of Lipschitz continuity?

    \item  Section~\ref{sec:effect_batch_size}: \textbf{Batch size}. What role does batch size play in shaping the temporal evolution of Lipschitz continuity?
      
    \item Section~\ref{sec:sampling_trajectory}: \textbf{Mini-batch sampling trajectory}. What impact does mini-batch sampling have on the temporal evolution of Lipschitz continuity?


\end{enumerate}

\begin{figure}[t]
    \centering

  \begin{minipage}[t]{0.48\textwidth}
    \begin{figure}[H]
      \centering
      \includegraphics[width=\linewidth]{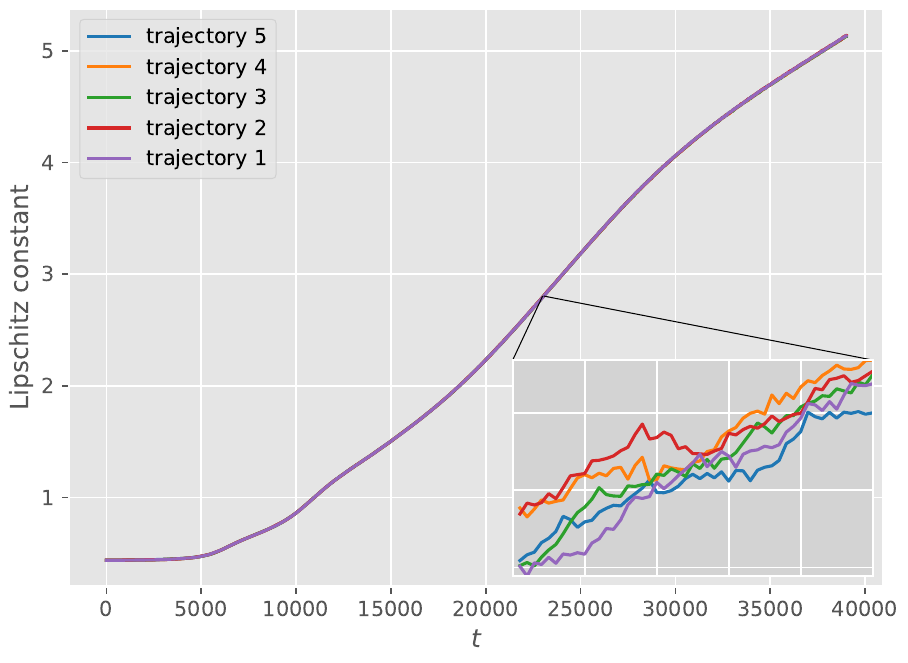}
      \caption[Predicted Effect of Trajectory in Mini-Batch Sampling on Lipschitz Continuity]{Predicted effect of the trajectory in mini-batch sampling on Lipschitz continuity. The inset plot zooms in on $50$ steps at $t=23000$.} 
      \label{fig:effect_trajectories}
    \end{figure}
  \end{minipage}
  \hfill
  \begin{minipage}[t]{0.48\textwidth}
    \begin{figure}[H]
      \centering
      \includegraphics[width=\linewidth]{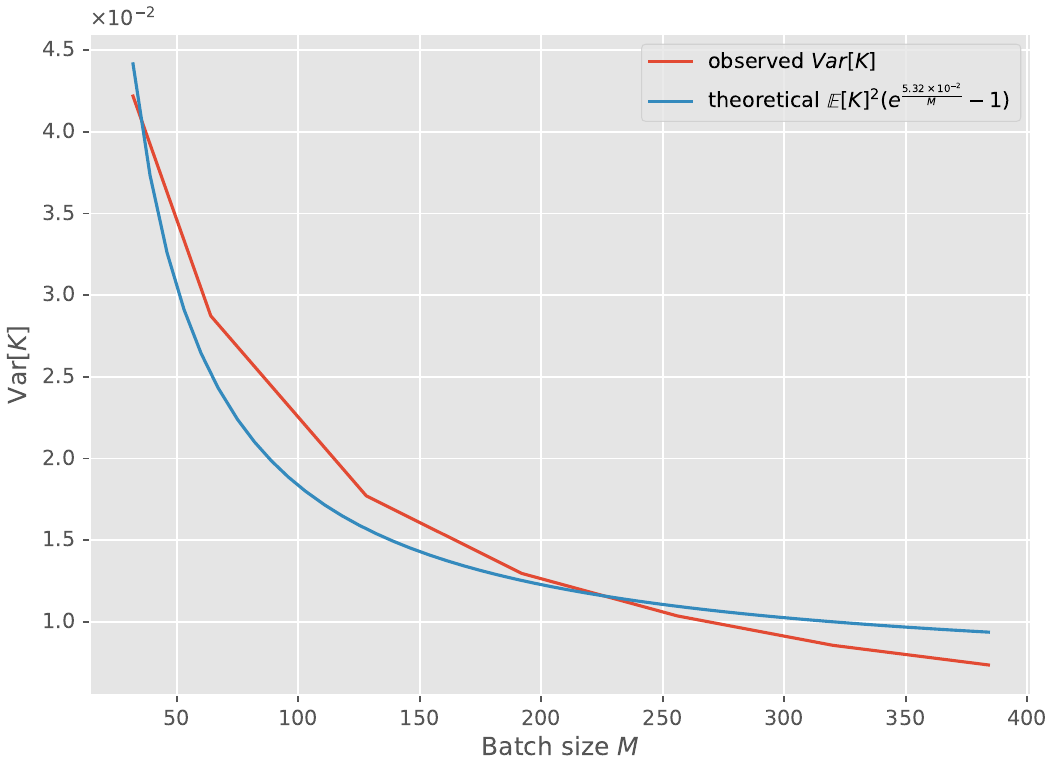}
      \caption[Predicted Effect of Batch Size on Variance of Lipschitz Continuity]{Predicted effect of batch size on the variance of Lipschitz continuity.}
      \label{fig:effect_batchsize}
    \end{figure}
  \end{minipage}

  \begin{minipage}[t]{0.48\textwidth}
    \begin{figure}[H]
      \centering
      \includegraphics[width=\linewidth]{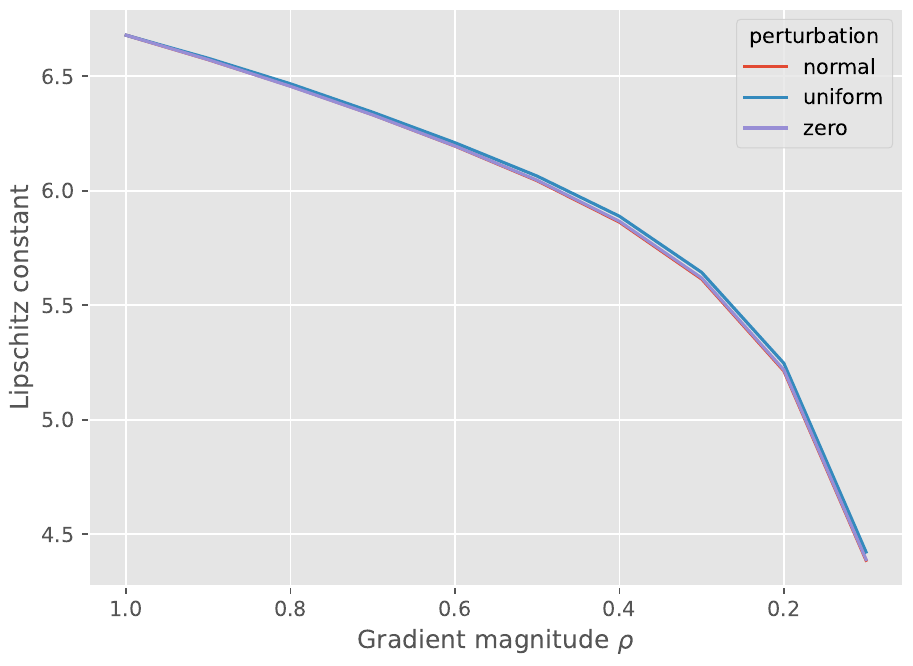}
      \caption[Predicted Effect of Gradient Magnitude and Perturbation]{Predicted effect of gradient magnitude and perturbation.}
      \label{fig:effect_gradmagnitude}
    \end{figure}
  \end{minipage}
  \hfill
  \begin{minipage}[t]{0.48\textwidth}
    \begin{figure}[H]
      \centering
      \includegraphics[width=\linewidth]{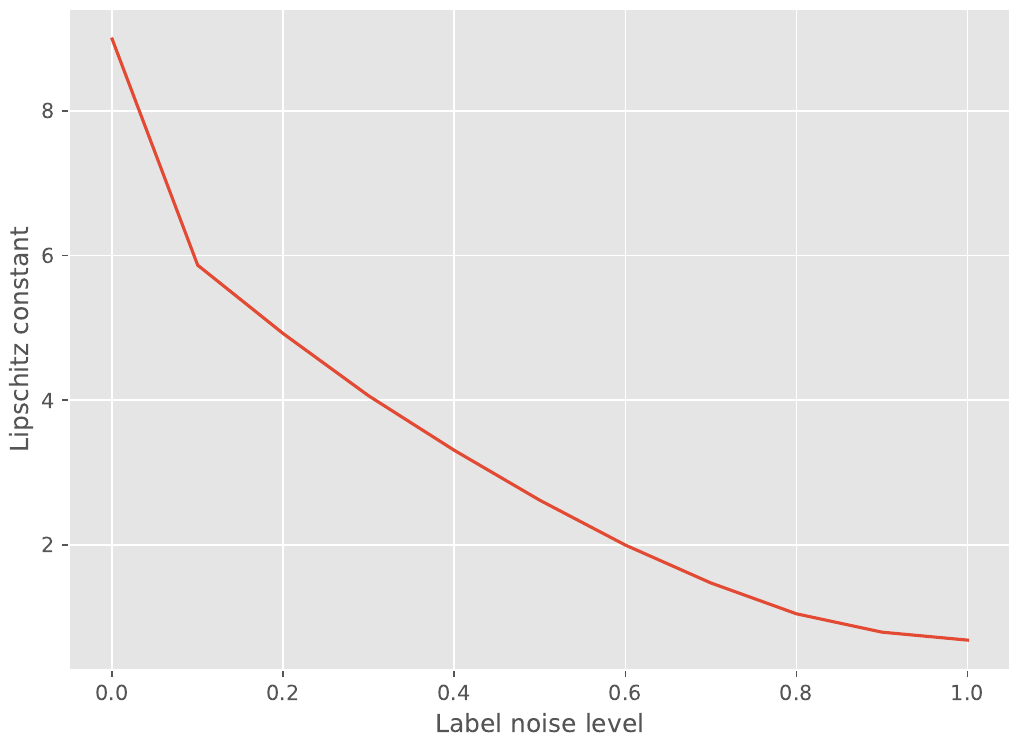}
      \caption[Predicted Effect of Uniform Label Corruption]{Predicted effect of uniform label corruption.}
      \label{fig:effect_labelnoise}
    \end{figure}
  \end{minipage}
  
\end{figure}

\subsection{Key factors driving dynamics}
\label{sec:key_factors}

\textbf{Principal driving factors}. Theorem~\ref{theorem:layer_dynamics} and Theorem~\ref{theorem:network_dynamics} identify four principal factors driving the evolution of Lipschitz continuity at both the layer and network levels: 
\begin{enumerate}[label=(\roman*)]
 
    \item \textbf{Gradient-principal-direction alignment}. The alignment between the gradient expectations and the principal directions of parameter matrices determines \textit{optimization-induced drift} the $\mu_Z(t)$. As shown in Figure~\ref{fig:longrange_dynamics}, this force dominates the drift of the Lipschitz continuity (see Theorem~\ref{theorem:layer_dynamics} and Theorem~\ref{theorem:network_dynamics}).

    \item \textbf{Directional gradient noise}. The directional projection of gradient fluctuations, arising from mini-batch sampling, on the principal directions of parameter matrices determines the \textit{diffusion-modulation intensity} $\lambda_Z(t)$ arises from. As shown in Figure~\ref{fig:longrange_dynamics}, this factor modulates the diffusion of the Lipschitz continuity.
    
    \item \textbf{Deterministic effect of gradient noise}. The interaction between the gradient fluctuations, arising from mini-batch sampling, and the operator-norm curvatures determines the \textit{noise-curvature entropy production} $\kappa_Z(t)$. Our theoretical framework suggests that this term gives a deterministic, non-negative, and irreversible increase contributing to the drift of the dynamics.

    \item \textbf{Gradient noise amplification effect}. Lemma~\ref{lemma:hessian_spectral_norm} and Theorem~\ref{theorem:layer_dynamics} suggest that the inverted principal spectral gaps in parameter matrices, defined as $\left[\sigma_1^{(\ell)}(t)^2 - \sigma_j^{(\ell)}(t)^2\right]^{-1}$, for all $j \neq 1$, can amplify the gradient noise. Thus smaller spectral gaps amplifies gradient noise, contributing to larger \textit{noise-curvature entropy production} $\kappa_Z(t)$.

\end{enumerate}

\subsection{Parameter initialization}
\label{sec:parameter_initialization}


According to Theorem~\ref{theorem:statistics_of_lipschitz}, the network initialization $K(0)$ affects both the expectation and uncertainty of Lipschitz continuity in the evolution. Suppose that the entries of a parameter $\theta^{(\ell)} \in \mathbb{R}^{m_{\ell} \times n_{\ell}}$ are sampled from $\mathcal{N}(0, s_{\ell}^2)$. According to the Tracy-Widom limiting distribution \citep{tracy1994level,tao2012topics}, the largest singular value limit is $\sigma_1^{\ell} \to (\sqrt{m_\ell} + \sqrt{n_\ell}) \cdot s_{\ell}$, if $m_\ell \cdot n_\ell$ is sufficiently large. For a network with $L$ layers, 
\begin{align}
K(0) \to \prod_{\ell=1}^{L}\left(\sqrt{m_{\ell}} + \sqrt{n_{\ell}}\right)  \cdot s_{\ell} \notag
     ,
\end{align}
hence a larger network has a larger Lipschitz continuity. For the case of Kaiming initialization, $s_{\ell} = \sqrt{\sfrac{2}{n_{\ell}}}$ \citep{he2015delving}.


\subsection{Unbounded growth near-convergence}
\label{sec:near_convergence}

Suppose a negative‐log‐likelihood loss:
\begin{align}
    \ell_f(\boldsymbol{\theta}(t); x, y) = -\log p_{\boldsymbol{\theta}(t)} (y\mid x) \notag
\end{align}
near convergence:
\begin{align}
   \ell_f(\boldsymbol{\theta}(t); x, y) \to 0 \notag
   . 
\end{align}

Under Assumption~\ref{assump:layerwise_noise}, 
the Fisher information matrix (FIM) of $p_{\boldsymbol{\theta}(t)} (y\mid x)$ with respect to parameters $\boldsymbol{\theta}^{(\ell)}(t)$ \citep{martens2020new} is defined as
\begin{align}
\label{equ:fim_approx_sigma}
    \boldsymbol{F}^{(\ell)}(t) &:= \mathbb{E} \left[
    \operatorname{vec}\left[
    \nabla^{(\ell)} p_{\boldsymbol{\theta}(t)} (y\mid x)\right] 
    \operatorname{vec}\left[
    \nabla^{(\ell)} p_{\boldsymbol{\theta}(t)} (y\mid x)\right]^\top
    \right] \approx M \cdot \boldsymbol{\Sigma}_t^{(\ell)}  \notag
    ,
\end{align}
as $\ell_f(\theta; x, y) \to 0$, where $M$ is the batch size \citep{jastrzkebski2017three,stephan2017stochastic,li2021happens,jastrzkebski2017three,martens2020new}. This will serve as a foundation for our later theoretical analysis of the near‐convergence dynamics.

\begin{proposition}[Unbounded Growth Near-Convergence]
Consider the terms in the dynamics (Theorem~\ref{theorem:layer_dynamics}):
\begin{gather}
\mu^{(\ell)}(t) =  \frac{1}{\sigma_{1}^{(\ell)}(t)} \left\langle \boldsymbol{J}_{op}^{(\ell)}(t), \, -\mathrm{vec}\left[ \nabla^{(\ell)} \mathcal{L}_f(\boldsymbol{\theta}(t)) \right] \right\rangle 
\approx 0 , \notag\\
\boldsymbol{\lambda}^{(\ell)}(t) = \frac{\sqrt{\eta}}{\sigma_{1}^{(\ell)}(t)}  
\left(\left[\boldsymbol{\Sigma}_t^{(\ell)}\right]^{\frac{1}{2}}\right)^\top 
\boldsymbol{J}_{op}^{(\ell)}(t)
 \approx 
 \frac{\sqrt{\eta}}{\sigma_{1}^{(\ell)}(t)}  
\left(\left[\frac{1}{M}\boldsymbol{F}^{(\ell)}(t)\right]^{\frac{1}{2}}\right)^\top 
\boldsymbol{J}_{op}^{(\ell)}(t) 
\to \boldsymbol{0}, \notag\\
\kappa^{(\ell)}(t) =\frac{\eta}{2\sigma_{1}^{(\ell)}(t)} \left\langle \boldsymbol{H}_{op}^{(\ell)}(t),  \boldsymbol{\Sigma}_t^{(\ell)}\right\rangle
\approx
\frac{\eta}{2\sigma_{1}^{(\ell)}(t)} \frac{1}{M}\left\langle \boldsymbol{H}_{op}^{(\ell)}(t),  \boldsymbol{F}^{(\ell)}(t)\right\rangle > 0
\notag
,
\end{gather}
so that Lipschitz continuity is not bounded in expectation:
\begin{align}
\lim_{t \to 0}
\mathbb{E}\left[K(t)\right]  
    = \lim_{t \to 0}\mathrm{e}^{Z(0) + \int_0^t \left[ \mu_Z(s) + \kappa_Z(s)  \right] \dd s}
    = 
    \lim_{t \to 0}
    \mathrm{e}^{Z(0) + \int_0^t \kappa_Z(s)  
    \dd s} \to \infty \notag
    ,
\end{align}
and uncertainty:
\begin{align}
\lim_{t \to 0}
\operatorname{Var}\left[K(t)\right] 
=
\lim_{t \to 0}
\mathbb{E}\left[K(t)\right]^2\left( e^{\operatorname{Var}\left[Z(t)\right]} - 1\right) \to \infty \notag
,
\end{align}
as $t \to \infty$. This result demonstrates that the Lipschitz continuity bound monotonically increases as optimization progresses. This phenomenon has been noted in literature \citep{yoshida2017spectral,bartlett2017spectrally,sedghi2019singular,gamba2023lipschitz}.

\end{proposition}

   
\medskip
\noindent
\textbf{Experimental Results}. We extend the training of the five-layer ConvNet on
CIFAR-10 without regularization to
$344,370$ steps ($1766$ epochs) for observing the long range dynamics near convergence. Figure~\ref{fig:dynamics:layer} shows the layer-specific dynamics; Figure~\ref{fig:dynamics:network} shows the network-specific dynamics. The observations are as follows:
\begin{enumerate}[label=(\roman*)]

    \item \textbf{Unbounded growth}. We train the network for up to $344,370$ steps, near convergence. The gradual increase driven by $\kappa_Z(t)$ persists throughout the training, with no observable indication of stopping, suggesting that the dynamics do not admit a finite-time steady state.

    \item \textbf{Non-negligible contribution of gradient noise}. Although the magnitude of $\kappa_Z(t)$ is typically one order smaller than that of $\mu_Z(t)$ in our experiments, its cumulative effect on the growth of Lipschitz continuity near-convergence is non-negligible and persists throughout training.

    \item \textbf{Four Phases}. We observe that $\kappa_Z(t)$ exhibits four distinct phases: (i) phase 1 --- a rapid initial increase at the beginning of training, (ii) phase 2 --- large fluctuations prior to overfitting, (iii) phase 3 --- a steady decline, and (iv) phase 4 --- convergence to a non-negative constant.
    
\end{enumerate}

\subsection{Noisy gradient regularization}
\label{sec:noisy_gradient}

Prior work shows that injecting noise into gradients during training can improve robustness \citep{laskey2017dart,welling2011bayesian} and help escape local minima \citep{neelakantan2015adding}. Our theoretical framework elucidates the mechanism by which gradient noise serves as a regularizer for Lipschitz continuity. Consider a loss $\ell_*$ with noisy supervision consists of a signal loss $\ell_f$ and a noisy loss $\ell_n$:
\begin{align}
\label{equ:noisy_loss}
    \ell_*(\boldsymbol{\theta};x,y) = \sqrt{\rho} \cdot  \ell_f(\boldsymbol{\theta};x,y) + 
    \sqrt{1-\rho} \cdot \ell_n(\boldsymbol{\theta};x,y) ,
\end{align}
where $\rho \in [0, 1]$. The corresponding batch loss is:
\begin{align}
    \nabla^{(\ell)} \mathcal{L}_*(\boldsymbol{\theta};\xi) = \sqrt{\rho} 
    \cdot \nabla^{(\ell)} \mathcal{L}_f(\boldsymbol{\theta};\xi) + 
    \sqrt{1-\rho} \cdot \nabla^{(\ell)} \mathcal{L}_n(\boldsymbol{\theta};\xi), \notag
\end{align}
and the gradient noise covariance is:
\begin{align}
    \Sigma^{(\ell),*} = \rho \cdot \Sigma^{(\ell),f} + (1-\rho) \cdot \Sigma^{(\ell),n} \notag
    .
\end{align}
If the noise is sampled from $\mathcal{N}(0, \sigma^2)$, $\Sigma^{(\ell),n} = \sigma^2\;\mathrm{I}$, the dynamics has the changes:
\begin{gather}
\mu_*^{(\ell)}(t) =  \sqrt{\rho} \mu^{(\ell)}(t)  \notag
, 
\end{gather}
and
\begin{align}
\kappa_*^{(\ell)}(t) &= \rho \cdot \kappa^{(\ell)}(t) + 
(1 -\rho) \cdot \frac{\eta}{2\sigma_{1}^{(\ell)}(t)} 
\underbrace{
\left\langle \boldsymbol{H}_{op}^{(\ell)}(t),  \sigma^2\; \mathrm{I} \right\rangle 
}_{\text{variance is negligible}}
\approx \rho \cdot \kappa^{(\ell)}(t) \notag
.
\end{align}
The variance contribution from $\boldsymbol{H}_{op}^{(\ell)}(t)$ is neglected, as experimental results indicate it is negligible (see Figure~\ref{fig:longrange_dynamics}). According to Theorem~\ref{theorem:statistics_of_lipschitz}, by ignoring $\lambda_Z(t)$, we have:
\begin{align}
\label{equ:gradient_magnitude}
\mathbb{E}\left[K(t)\right]  
    \approx \mathrm{e}^{Z(0) + \int_0^t  \sqrt{\rho} \cdot \mu_Z(s)  \dd s + \int_0^t  \rho \cdot \kappa_Z(s)  \dd s}
    = 
    K(0)\left[ \mathrm{e}^{\int_0^t  \mu_Z(s)  \dd s} \right]^{\sqrt{\rho}} \left[ \mathrm{e}^{\int_0^t  \kappa_Z(s)  \dd s} \right]^{\rho}
    .
\end{align}
Although the integral $\int_0^t \mu_{Z}(s) \dd s$ and $\int_0^t \kappa_{Z}(s) \dd s$ are unlikely to be computed, it provides insight how gradient magnitude affects the evolution of Lipschitz continuity bound. This result shows higher gradient magnitude leads larger Lipschitz continuity bound. This framework can be used to analyze the uniform label corruption (see Section~\ref{sec:uniform_label_noise}).

\medskip
\noindent
\textbf{Experimental Results}. We train a MLP (see Table~\ref{tab:network_config_mlp}) on the MNIST with a batch size of $128$ and a learning rate of $0.01$, using SGD without momentum. During training, the gradients are scaled by $\sqrt{\rho}$, and additive noise is injected sampled from: (i) a Gaussian distribution $\mathcal{N}(0, 1)$; (ii) a uniform distribution $U[-0.5, 0.5]$; and (iii) zero noise, scaled by $\sqrt{1 - \rho}$. As shown in Figure~\ref{fig:effect_gradmagnitude}, the Lipschitz constant decreases monotonically as $\rho$ decreases.

\subsection{Uniform label corruption}
\label{sec:uniform_label_noise}
Suppose a classifier with a negative-log-likelihood loss function $\ell_f(\theta;x,y):= -\log p_{\theta}(y|x)$. Further suppose that the label corruption probability is $\epsilon \in [0, 1]$ with a uniform distribution $U(\mathcal{Y})$, while the label remains intact with a probability $1-\epsilon$. Therefore the gradient with label corruption can be expressed as \citep{natarajan2013learning,sukhbaatar2014training,patrini2017making,ghosh2017robust}:
\begin{align}
 \mathbb{E}\left[
 \nabla^{(\ell)} 
 \ell_f(\theta;x,y)
 \right]
 := 
 (1-\epsilon)\cdot 
 \mathbb{E}_{x,y}\left[\nabla^{(\ell)} 
 \left[-\log p_{\theta}(y|x)\right] 
 \right]
 + 
 \epsilon \cdot 
 \mathbb{E}_x\left[
 \nabla^{(\ell)}
 \frac{1}{|\mathcal{Y}|}\sum_{ \Tilde{y} \sim U(\mathcal{Y}) }
 \left[-\log p_{\theta}(\Tilde{y}|x)\right]
 \right]
 ,
 \notag
\end{align}
where $\Tilde{y} \sim U(\mathcal{Y}) $. Literature \citep{natarajan2013learning,sukhbaatar2014training,ghosh2017robust} assume that the loss component with label noise contributes only a constant, hence zero gradient:
\begin{align}
\langle 
\boldsymbol{J}_{op}^{(\ell)}(t), \operatorname{vec}\left[ -\nabla^{(\ell)} \mathbb{E}_{x,\Tilde{y}}\left[ -\log p_{\theta}(\Tilde{y}|x) \right]\right]
\rangle
\to 0. \notag
\end{align}

According to Equation~\ref{equ:noisy_loss}, we can treat the label noise as $\ell_n$. By using Equation~\ref{equ:gradient_magnitude}, note $\rho = (1-\epsilon)^2$, then we have:
\begin{align}
\mathbb{E}\left[ K'(t)\right] \approx K(0) \cdot \left[ \mathrm{e}^{\int_0^t \mu_Z(s) \dd s} \right]^{1-\epsilon} 
\cdot \left[ \mathrm{e}^{\int_0^t \kappa_Z(s) \dd s} \right]^{(1-\epsilon)^2}
\notag
.
\end{align}


\medskip
\noindent
\textbf{Experimental Results}. We train a MLP (see Table~\ref{tab:network_config_mlp}) on the MNIST with a batch size of $128$ and a learning rate of $0.01$, using SGD without momentum. We inject uniform label noise into the dataset with a level from $0$ to $1$. As shown in Figure~\ref{fig:effect_labelnoise}, the Lipschitz constant decreases monotonically as label noise level increases.

\subsection{Batch size}
\label{sec:effect_batch_size}

Suppose that the batch size $M$ in mini-batch sampling affects only the gradient noise $\Sigma(t)$, and does not affect the expectations of gradients $\nabla \mathcal{L}_f(\theta)$. Therefore, the batch size $M$ affects only the diffusion term (Theorem~\ref{theorem:statistics_of_lipschitz}). We analyze the effect of mini-batch size.

\begin{proposition}[Uncertainty Under Large Batch]
\label{proposition:effect_batch_size}
Let the batch size be $M$ and sufficiently large, then:
\begin{align}
\mathrm{Var}\left[ K(t) \right] 
&\approx \mathbb{E}\left[K(t)\right]^2 \frac{C}{M} ,\notag
\end{align}
where $C$ is a constant given as:
\begin{align}
C=
\displaystyle\int_0^t \sum_{l=1}^{L}  \frac{\eta}{\sigma_1^{(\ell)} (t)^2} \boldsymbol{J}_{op}^{(\ell)}(t)  \mathrm{Var}\left[ \mathrm{vec}\left(\nabla^{(\ell)} \ell_f(\boldsymbol{\theta}(t); x, y) \right)\right]\boldsymbol{J}_{op}^{(\ell)}(t)^\top \, \dd s     \notag
\end{align}
\end{proposition}
\begin{proof}
    
Substitute Equation~\ref{equ:batch_gradient_noise} into Theorem~\ref{theorem:statistics_of_lipschitz}:
\begin{align}
    \mathrm{Var}\left[ Z(t) \right] &= \displaystyle\int_0^t \lambda_Z(s)^2 \, \dd s = \displaystyle\int_0^t \sum_{\ell=1}^{L} \|\boldsymbol{\lambda}^{(\ell)}(s)\|_2^2 \, \dd s = \displaystyle\int_0^t \sum_{l=1}^{L}  \frac{\eta}{\sigma_1^{(\ell)}(t)^2} \boldsymbol{J}_{op}^{(\ell)}(t) \boldsymbol{\Sigma}^{(\ell)}(t) \boldsymbol{J}_{op}^{(\ell)}(t)^\top \, \dd s \notag \\
    &= \displaystyle\int_0^t \sum_{l=1}^{L}  \frac{\eta}{\sigma_1^{(\ell)}(t)^2} \boldsymbol{J}_{op}^{(\ell)}(t)  
    \frac{1}{M} \mathrm{Var}\left[ \mathrm{vec}\left(\nabla^{(\ell)} \ell_f(\boldsymbol{\theta}(t); x, y) \right)\right]
    \boldsymbol{J}_{op}^{(\ell)}(t)^\top \, \dd s \notag\\
    &= \frac{1}{M} 
    \underbrace{
    \displaystyle\int_0^t \sum_{l=1}^{L}  \frac{\eta}{\sigma_1^{(\ell)} (t)^2} \boldsymbol{J}_{op}^{(\ell)}(t)  \mathrm{Var}\left[ \mathrm{vec}\left(\nabla^{(\ell)} \ell_f(\boldsymbol{\theta}(t); x, y) \right)\right]\boldsymbol{J}_{op}^{(\ell)}(t)^\top \, \dd s 
    }_{\text{a constant}}
     =  \frac{1}{M} C . \notag
\end{align}

Hence:
\begin{align}
\mathrm{Var}\left[ K(t) \right] &= \mathbb{E}\left[K(t)\right]^2\left(\mathrm{e}^{\mathrm{Var}\left[ Z(t) \right]} - 1\right) =\mathbb{E}\left[K(t)\right]^2\left(\mathrm{e}^{\frac{C}{M}} - 1\right) \notag
.
\end{align}

Suppose $M \to \infty$:
\begin{align}
\mathrm{Var}\left[ K(t) \right] &=\mathbb{E}\left[K(t)\right]^2\left(\mathrm{e}^{\frac{C}{M}} - 1\right) \approx \mathbb{E}\left[K(t)\right]^2 \left(1 + \frac{C}{M} - 1\right)  =\mathbb{E}\left[K(t)\right]^2 \frac{C}{M} 
\approxprop \frac{1}{M} \notag
.
\end{align}

\end{proof}

\medskip
\noindent
\textbf{Experimental Results}. We train the five-layer ConvNet (Table~\ref{tab:network_config_cnn} in the Appendix), optimized with SGD without momentum. The batch size is varied from $32$ to $384$ in increments of $32$, and each configuration is trained for $39{,}000$ steps. During training, we profile and collect the dynamics to compute the uncertainty of Lipschitz continuity $\mathrm{Var}\left[K(t)\right]$. As shown in Figure~\ref{fig:effect_batchsize}, the observed effect of batch size aligns closely with the theoretical prediction in Proposition~\ref{proposition:effect_batch_size}.

\subsection{Mini-batch sampling trajectory}
\label{sec:sampling_trajectory}

According to Proposition~\ref{proposition:effect_batch_size}, if the batch size is sufficiently large, the effect from various mini-batch sampling trajectories can be neglected since:
\begin{align}
\mathrm{Var}\left[ K(t) \right] \approx \mathbb{E}\left[K(t)\right]^2 \frac{C}{M} 
 \to 0 \notag
,
\end{align}
as batch size $M$ is sufficiently large.

\medskip
\noindent
\textbf{Experimental Results}. We train the five-layer ConvNet (Table~\ref{tab:network_config_cnn} in the Appendix), optimized with SGD without momentum. The parameter initialization random seed is fixed to $1$, while the random seed for mini-batch sampling is varied from $1$ to $5$. As shown in Figure~\ref{fig:effect_trajectories}, the sampling trajectories have a negligible effect on the evolution of Lipschitz continuity.

\section{Limitations and future work}

While our mathematical framework shows strong agreement with empirical results, it has several limitations:
\begin{enumerate}
    \item \textbf{Layer-wise noise assumption}. The simplification in Assumption~\ref{assump:layerwise_noise} neglects inter-layer interactions among neurons; this may pose as a challenging for studying large models.
    
    \item \textbf{Distributional assumption}. We assume that the gradient noise follows a time-state-dependent normal distribution varying smoothly. However, this assumption may not hold for small batch sizes, where the noise distribution can deviate significantly from Gaussian.
    
    \item \textbf{Continuous–discrete error}. Our framework models gradient dynamics in continuous time, whereas SGD operates in discrete steps. This discrepancy may introduce non-negligible errors, particularly when modeling long-range dynamics.
\end{enumerate}

\section{Conclusions}

We present a mathematical framework that models the dynamics of Lipschitz continuity in neural networks through a system of SDEs. This theoretical framework not only identifies the key factors governing the evolution of Lipschitz continuity, but also provides insight into how it is implicitly regularized during training. Beyond its analytical value, the framework offers a foundation for future research and development. Our experiments further validate the effectiveness of the proposed framework and the predicted effects of the theoretical implications.




\acks{This publication has emanated from research conducted with the financial support of \textbf{Taighde \'Eireann} - Research Ireland under Grant number 18/CRT/6223. We extend heartfelt thank to Prof. Dr. Changjian Shui (University of Ottawa, Canada) for the constructive comments and valuable assistance. 
}


\newpage

\appendix

\section{Network configuration}

\begin{table}[h]
    \centering
    \adjustbox{max width=\textwidth}{%
    \begin{tabular}{|c|l|l|}
        \hline
        
        \textbf{Function Block} &\textbf{\# of Layer} &  \textbf{Module} \\
        \hline
        
        \multirow{3}{*}{Block \#1} & \cellcolor{lime!20}$1$ & \cellcolor{lime!20}\texttt{nn.Conv2d(3, 32, kernel\_size=3, padding=1)} \\
        \cline{2-3}
        
        &$2$ & \texttt{nn.ReLU()} \\
        \cline{2-3}

        &$3$ & \texttt{nn.MaxPool2d(kernel\_size=2, stride=2)} \\
        \hline

        \multirow{4}{*}{Block \#2}& \cellcolor{lime!20}$4$ &\cellcolor{lime!20}\texttt{nn.Conv2d(32, 64, kernel\_size=3, padding=1)} \\
        \cline{2-3}

        &$5$ (*) & \texttt{nn.Dropout(p=0.2)} \\
        \cline{2-3}

        &$6$ & \texttt{nn.ReLU()} \\
        \cline{2-3}

        &$7$ & \texttt{nn.MaxPool2d(kernel\_size=2, stride=2)} \\
        \hline

        \multirow{4}{*}{Block \#3}& \cellcolor{lime!20}$8$ &\cellcolor{lime!20}\texttt{nn.Conv2d(64, 128, kernel\_size=3, padding=1, stride=2)} \\
        \cline{2-3}

        &$9$ (*) & \texttt{nn.Dropout(p=0.2)} \\
        \cline{2-3}

        &$10$ (*) & \texttt{nn.BatchNorm2d(128, track\_running\_stats=False)} \\
        \cline{2-3}

        &$11$& \texttt{nn.ReLU()} \\
        \hline

        \multirow{4}{*}{Classifier} & \cellcolor{lime!20}$12$ &\cellcolor{lime!20}\texttt{nn.Linear(128 * 4 * 4, 256)} \\
        \cline{2-3}

        &$13$ & \texttt{nn.ReLU()} \\
        \cline{2-3}

        &$14$ (*) & \texttt{nn.Dropout(p=0.3)} \\
        \cline{2-3}

        &\cellcolor{lime!20}$15$ & \cellcolor{lime!20}\texttt{nn.Linear(256, num\_classes)} \\
        \hline
        
    \end{tabular}
    }
    \caption{ConvNet configuration. The layer marked as `(*)' is configurable with respect to experimental needs. There are five parameterized layers (\ie~\# 1, \# 4, \# 8, \# 12 and \# 15).}
    
    \label{tab:network_config_cnn}
\end{table}

\begin{table}[h]
    \centering
    \begin{tabular}{|l|l|}
        \hline
        
        \textbf{\# of Layer} &  \textbf{Module} \\
        \hline
        
        \cellcolor{lime!20}$1$ & \cellcolor{lime!20}\texttt{nn.Linear(28*28, 512)} \\
        \hline
        
        $2$ & \texttt{nn.ReLU()} \\
        \hline

         \cellcolor{lime!20}$3$ & \cellcolor{lime!20}\texttt{nn.Linear(512, 256)} \\
        \hline

        $4$ & \texttt{nn.ReLU()} \\
        \hline

        \cellcolor{lime!20}$5$ & \cellcolor{lime!20}\texttt{nn.Linear(256, 10)} \\
        \hline
        
    \end{tabular}
    \caption{MLP configuration. There are three parameterized layers (\ie~\# 1, \# 3, and \# 5).}
    
    \label{tab:network_config_mlp}
\end{table}

\section{Regularization configuration}

\begin{table}[h]
    \centering
    \adjustbox{max width=\textwidth}{%
    \begin{tabular}{|c|l|}
        \hline
        
        \textbf{Regularization} & \textbf{Configuration} \\
        \hline

        mixup & $\alpha=0.4$ \\
        \hline

        label smoothing & $\epsilon=0.1$ \\
        \hline
        
        adversarial training w/ FGSM & $\epsilon=0.03$ \\
        \hline
  
        weight-decay & $\lambda=0.001$ \\
        \hline

        \multirow{2}{*}{auto-augment} & \texttt{AutoAugment(policy=AutoAugmentPolicy.CIFAR10)} \\
        \cline{2-2}
        
        & \texttt{AutoAugment(policy=AutoAugmentPolicy.CIFAR100)} \\
        \hline
        
    \end{tabular}
    }
    \caption{Regularization configuration.}
    
    \label{tab:regularization_config}
\end{table}

\section{Full validation experiment}
\label{sec:full_validation_cifar10}

\begin{figure}[h]
  \centering

    \includegraphics[width=1\linewidth]{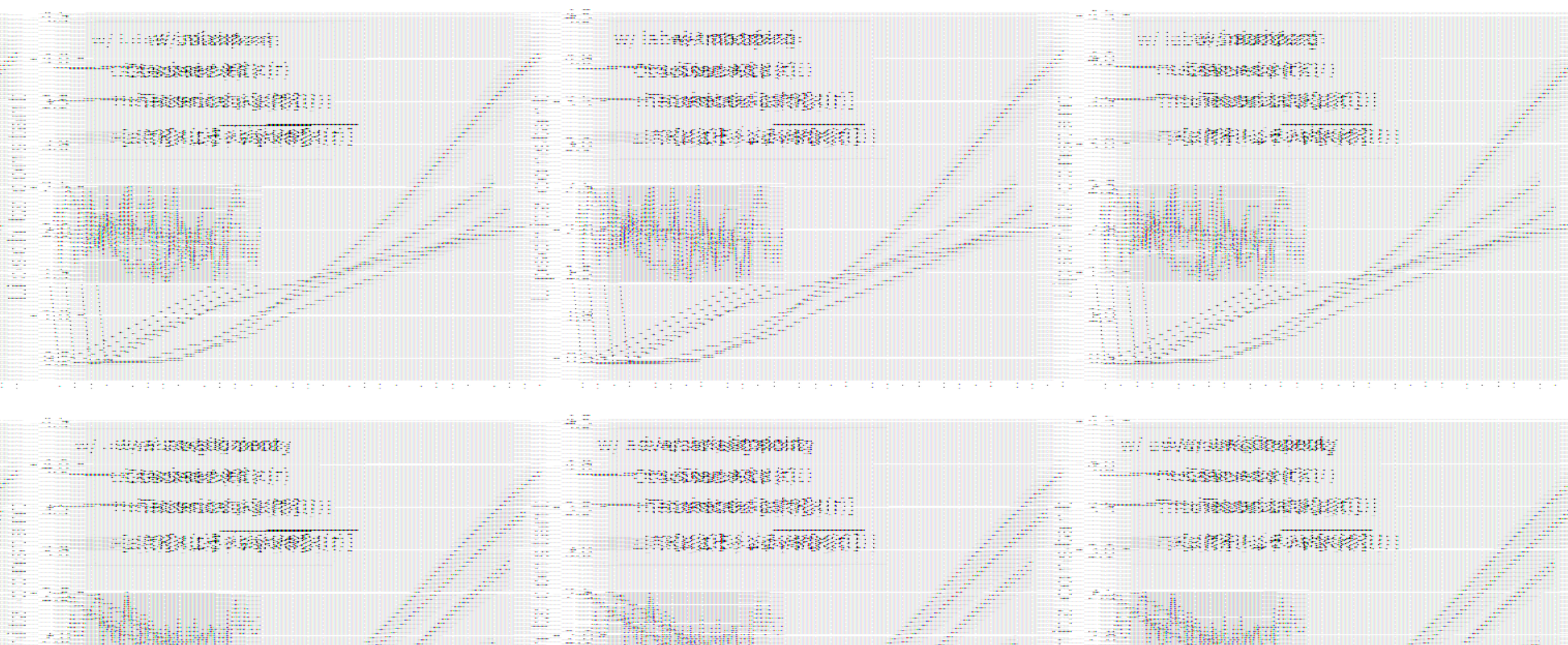}

   \caption[Full Numerical Validation of Our Mathematical Framework on CIFAR-10]{Full numerical validation of our mathematical framework on CIFAR-10.}
  
   \label{fig:validation:cifar10:full}
  
\end{figure}

Results of full validation experiments are shown as in Figure~\ref{fig:validation:cifar10:full}.

\vskip 0.2in
\bibliography{references}

\end{document}